%% file: example_paper.tex
\documentclass{article}

\usepackage{microtype}
\usepackage{graphicx}
\usepackage{subfigure}
\usepackage{booktabs} 

\PassOptionsToPackage{hyphens}{url}\usepackage{hyperref}

\usepackage[accepted]{icml2025}

\usepackage{amsmath}
\usepackage{amssymb}
\usepackage{mathtools}
\usepackage{amsthm}

\usepackage[capitalize,noabbrev]{cleveref}

\theoremstyle{plain}

\theoremstyle{definition}

\theoremstyle{remark}

\usepackage[disable,textsize=tiny]{todonotes}

\usepackage{paralist}
\input{common}

\usepackage{makecell}
\usepackage{bbding}
\usepackage{svg}

\icmltitlerunning{Identifying Sensitive Weights via Post-quantization Integral}

\begin{document}

\twocolumn[
\icmltitle{Identifying Sensitive Weights via Post-quantization Integral}



\icmlsetsymbol{equal}{*}

\begin{icmlauthorlist}
\icmlauthor{Yuezhou Hu}{thu}
\icmlauthor{Weiyu Huang}{thu}
\icmlauthor{Zichen Liang}{thu}
\icmlauthor{Chang Chen}{thu}
\icmlauthor{Jintao Zhang}{thu}
\icmlauthor{Jun Zhu}{thu}
\icmlauthor{Jianfei Chen}{thu}
\end{icmlauthorlist}

\icmlaffiliation{thu}{Dept. of Comp. Sci. \& Tech., Institute for AI, BNRist Center, Tsinghua-Bosch Joint ML Center, THBI Lab, Tsinghua University}

\icmlcorrespondingauthor{Jianfei Chen}{jianfeic@tsinghua.edu.cn}

\icmlkeywords{Efficient Machine Learning, Large Language Model, Quantization, Post-training Quantization}

\vskip 0.3in
]



\printAffiliationsAndNotice{\icmlEqualContribution} 

\begin{abstract}

Serving Large Language Models (LLMs) is costly. However, post-training weight quantization can address this problem by both compressing their sizes for limited memory and saving bandwidth for acceleration. As not all weight dimensions are equally important, those methods typically rely on a sensitivity metric, which indicates the element-wise influence of weights on loss function and is used to preprocess original weights for better quantization. In this work, we conduct an empirical study on the accuracy of the sensitivity metric, and find that existing gradient and Hessian based metrics are very inaccurate: they underestimate quantization's impact on the loss function by orders of magnitude, mainly due to the small convergence radius of local 2nd order approximation, \ie, gradient and Hessian term in Taylor's formula. To tackle this problem, we propose Post-quantization Integral (PQI), an accurate metric to estimate posterior sensitivity in a fine-grained manner. To leverage this accurate metric, we further propose ReQuant, a simple yet powerful framework that mainly consists of two Dense-and-Sparse detach components: self-adaptive outlier selection and step-wise significant weights detach. Results show that ReQuant boosts state-of-the-art post-training quantization methods, with a pronounced improvement of 2.66 perplexity gain on Llama 3.2 1B with QTIP.

\end{abstract}

\section{Introduction}
The past decade has witnessed the thriving of Large Language Models, which have exhibited their great potential in various domains, such as reasoning \cite{NEURIPS2022_9d560961}, code generation \cite{hui2024qwen25codertechnicalreport} and instruction following \cite{ouyang2022traininglanguagemodelsfollow}. However, as LLMs are becoming larger, efficiently serving or even simply launching them becomes a challenge, especially for those edge devices with limited memory. 
One possible solution to this is weight quantization, which converts a high-precision LLM into a low-precision counterpart. 
As the main bottleneck of LLM decoding is loading weights from memory to chip, weight quantization also proportionally accelerate decoding. 

Many previous methods have demonstrated the possibility to compress LLM into lower precision. Specifically, post-training quantization (PTQ) methods~\cite{frantar2023gptqaccurateposttrainingquantization,MLSYS2024_42a452cb,pmlr-v202-xiao23c,NEURIPS2023_0df38cd1,pmlr-v235-kim24f,lin2024duquant,tseng2024qtip,zhang2025sageattention,zhang2024sageattention2} are popular since they do not require the expensive training procedure.
Although most PTQ methods are feasible with models such as OPT \cite{zhang2022optopenpretrainedtransformer} and Llama 2 \cite{touvron2023llama2openfoundation}, the effectiveness of these methods on state-of-the-art models in preserving accuracy remains an open question. Empirical study~\cite{kumar2024scalinglawsprecision} reveals that the quantization difficulty depends on the ratio of dataset size  and model size. As recent LLMs, such as Phi-4~\cite{abdin2024phi4technicalreport}, are typically trained with more than ten trillion tokens, it is even challenging to compress them to 4-bit without accuracy degradation.

Existing low-bit quantization methods for LLMs exploit the unequal importance of weight dimensions, necessitating the use of a sensitivity metric to quantify their impact. Extensive experiments have demonstrated that certain dimensions are more critical than others, where even slight modifications to these weights can cause significant distortions in the model’s output~\cite{MLSYS2024_42a452cb}. Therefore, an accurate \emph{sensitivity metric} for weight dimensions is essential for the effectiveness of a quantization algorithm.
Consider quantizing a model $\mathcal{M}$ with weights $\wv\in \mathbb{R}^D$ using a calibration set $\Dc$. 
A PTQ method typically consists of two stages. In the first stage, a sensitivity metric $S$ is calculated on the calibration set, which is typically a vector indicating the element-wise\footnote{For channel-wise sensitivity metrics such as AWQ \cite{MLSYS2024_42a452cb}, the definition of sensitivity can also be extended to be element-wise by simple broadcasting along the other dimension.} importance: $\vv=S(\mathcal{M},\Dc)\in \mathbb{R}^D$. 
In the second stage, the sensitivity vector is used to quantize $\wv$: $\Tilde{\wv} = Q(\wv,\vv)$, where $Q$ is the quantization process, and $\Tilde{\wv}\in \mathbb{R}^D$ is the low-precision weight vector. For example, the simplest technique is to store the most sensitive elements in full precision to preserve accuracy. For another, a scale can be applied to salient weights to diminish the relative quantization error.

In this work, we empirically evaluate the accuracy of sensitivity metrics in predicting the change in the loss function caused by weight quantization. Unfortunately, \emph{none of the existing sensitivity metrics is sufficiently accurate}, in the sense that they cannot predict the change of loss function caused by weight quantization.  We identify two key reasons why previous methods fail to accurately estimate sensitivity:
1) \textbf{Small convergence radius:} The complicated loss landscape of LLM makes the local gradient and Hessian based approximation only valid in a very small region near $\wv$. Specifically, the quantized weight $\Tilde{\wv}$ falls outside the convergence radius. 
2) \textbf{Misalignment of Sensitivity between $\wv$ and $\Tilde{\wv}$:} While previous methods only calculate the sensitivity based on $\wv$, sensitivity results might change: previously sensitive weights may lose importance after quantization, and other non-sensitive weights may emerge as sensitive.

To overcome these difficulties, we introduce PQI, a novel sensitivity metric to estimate the influence of each quantized weight. First, PQI only leverages the local continuity of the model, which can be easily achieved. Additionally, both $\wv$ and $\Tilde{\wv}$ are considered to calculate $\vv$, making it more accurate.
Our contributions are summarized as follows:
\begin{compactitem}[$\bullet$]
\item We propose Post-quantization Integral (PQI), an accurate sensitivity metric designed to compute the element-wise importance of a quantized model.
\item We introduce ReQuant, a pipeline that utilizes PQI to enhance the quality of a quantized model by a Dense-and-Sparse  \cite{li2023losparsestructuredcompressionlarge,pmlr-v235-kim24f} decomposition.
\item When applied to Llama 3.2 1B model, ReQuant can reduce the perplexity up to 2.6 and enhance MATH \cite{NEURIPS_DATASETS_AND_BENCHMARKS2021_be83ab3e} few-shot by nearly 3\%, compared to SqueezeLLM \cite{pmlr-v235-kim24f} and QTIP \cite{tseng2024qtip} baselines.
\end{compactitem}


\section{Related Work}
\paragraph{Post-training Quantization}
Early attempts to quantize transformers \cite{NEURIPS2022_1caf09c9,frantar2023gptqaccurateposttrainingquantization} utilize Hessian to calibrate quantized weights. Other studies focus on activation outliers \cite{NEURIPS2022_c3ba4962,pmlr-v202-xiao23c,MLSYS2024_42a452cb}. However, all of these studies either suffer from suboptimal accuracy or low compression ratio problems.
Another method is to resolve outliers via rotation \cite{ma2024affinequant,shao2024omniquant,lin2024duquant,ashkboos2024quarotoutlierfree4bitinference,liu2024spinquantllmquantizationlearned}. However, those methods require a thorough search of the rotation matrices, which brings extra costs. Codebook-based algorithms \cite{tseng2024quip,pmlr-v235-kim24f,tseng2024qtip} compress the model into a pre-computed lookup table and its indices. Other lookup-free methods \cite{NEURIPS2023_1feb8787} can also be categorized as a pre-established codebook. Constructing the codebook is non-trivial, since the distribution of the codebook needs to be aligned with weight distributions. Thus, most of the intuitive codebooks can only reach suboptimal performance.

\paragraph{Other Quantization Methods} Quantizaion Aware Training (QAT) methods dynamically quantize weights during training rather than after training \cite{liu-etal-2024-llm,wang20241bitaiinfra11}. However, this technique is still immature and requires considerable training cost. 
In contrast to accuracy-centric studies, others focus on activation and KV cache quantization \cite{liu2023llmqatdatafreequantizationaware,hooper2024kvquant,lin2024qserve}, which strikes a balance between accuracy and fast inference. Since outliers in activations are equally or even more important than in weights, those methods typically adopt naive grouping-based quantization, which can maximize inference speed with higher accuracy loss than post-training quantization methods. Some other studies investigate data parallelism \cite{jia2024sdpbit},  adapters for quantized models \cite{NEURIPS2023_1feb8787,li2024loftq,xia2024efficient}, and low-bit optimizers \cite{dettmers2022bit,NEURIPS2023_3122aaa2}, etc. These studies involve model pre-training and are orthogonal to our study.

\section{Background}
In this section, we first present an overview of PTQ methods. Afterward, we revisit common sensitivity metrics in them.

\subsection{Post-training Quantization}
We classify PTQ methods into two primary categories based on their quantization strategy.
\paragraph{Grouping-based Quantization}
Let $\{\wv_i\}_{1\leq i \leq d}$ be $d$ group splits of $\wv$, and corresponding sensitivity $\{\vv_i\}_{1\leq i \leq d}$. Each group is quantized with its own quantization step $s_i$:
\begin{align*}
    \hat{\wv}_{i}&=\operatorname{preprocess}(\wv_i,\vv_i),\\
    \Tilde{\wv}_i&=s_i\cdot\operatorname{round2int}(\frac{\hat{\wv}_{i}}{s_i}),s_i=\frac{\max\abs{\wv_i}}{2^N-1},
\end{align*}
where $\operatorname{preprocess}$ is basically a transformation that makes sensitive elements more accurate to quantize, and $\operatorname{round2int}$ performs element-wise rounding. For example, in GPTQ \cite{frantar2023gptqaccurateposttrainingquantization}, the preprocessing step is performed by dynamically adjusting Hessian and quantized weights, and in AWQ, by multiplying channel-wise activations.

\paragraph{Codebook-based Quantization} Methods based on codebooks relie on a pre-computed lookup table $\Tv\in\mathbb{R}^K$, which is generally a vector of $K$ possible high-precision values built from $\wv$ and $\vv$, and $\Tilde{\wv}$ is obtained by rounding $\wv$ to the nearest possible entry in $\Tv$:
\begin{align*}
    \Tv&=\operatorname{build}(\wv,\vv)\\
    \Tilde{\wv}&=\operatorname{round2nearest}(\wv,\Tv).
\end{align*}
For example, in SqueezeLLM, $\Tv$ is with K-means: $\min_{\tv}(\tv-\wv)^\top\Hv(\tv-\wv)$, with the clustering center vector $\tv$ serving as entries in $\Tv$. $\Hv\in\mathbb{R}^{D\times D}$ is the diagonal Hessian matrix and $\vv$ is the non-zero elements in it:
$\vv=\operatorname{diag}(\Hv)$.
In QTIP, the table is constructed based on Gaussian distribution. Note that some methods such as QLoRA \cite{NEURIPS2023_1feb8787} do not require a sensitivity metric, and build the lookup table merely on weight distributions.

\subsection{Sensitivity Metrics}
The sensitivity $\vv$ is the metric used to signify the importance of each weight.  Of note, we use $f$ to denote the loss function, $F$ to denote the optimization target: $F(\wv)=\frac{1}{n} \sum \limits_{i=1}^n f (\wv;\xv_{[i]})
$, with $\Dc = \{\xv_{[i]}\}_{i=1}^n$ denoting a calibration set of $n$ samples. An ideal sensitivity metric should accurately predict the change of loss function $F(\Tilde{\wv})-F(\wv)$, which we denote as $\Delta F$.
Theoretically, under ideal conditions with $K$th-order continuously differentiable partial derivatives around $\mathbf{w}$, $\Delta F$ can be expressed with Taylor series:
\begin{align*}
    \Delta F=\sum_{k=1}^{K}\frac{1}{k!}\left(\sum_{i=1}^{n}(\Tilde{w}_i-w_i)\frac{\partial}{\partial w_i}\right)^kF(\wv)+o(\abs{\Tilde{\wv}-\wv}^K).
\end{align*}
By keeping the first two terms, $\Delta F$ can be reduced as:
\begin{small}
\begin{align}
\label{eqn:first-and-second}
    \Delta F\approx\nabla_{\wv} F(\wv)^\top(\Tilde{\wv}-\wv)+\frac{1}{2}(\Tilde{\wv}-\wv)^\top\Hv(\Tilde{\wv}-\wv).
\end{align}
\end{small}
Taking the special cases of Eq. (\ref{eqn:first-and-second}) derives three metrics that are widely used: gradient, activation, and Hessian. It is worth noticing that the three metrics are calculated from the dense model  before quantization. This means that we can only make use of the original weights $\wv$ to predict loss function change.
\begin{compactitem}[$\bullet$]
    \item \textbf{Gradient:} \citet{shao2024gwqgradientawareweightquantization} propose to use gradient as sensitivity: $\vv=\nabla_{\wv}F(\wv)$, where they choose to keep weights with the largest gradients in 16 bits. This approach retains the first-order term in Taylor's formula:
    \begin{align}
    \label{eqn:gradiet}
        \Delta F\approx\nabla_{\wv}F(\wv)^\top(\Tilde{\wv}-\wv).
    \end{align}
    \item \textbf{Activation:} \citet{MLSYS2024_42a452cb} propose to use input activation of each linear layer as its sensitivity. Note that activation equals gradient only in simple cases (one linear layer with L1 norm loss). Since a LLM consists of multiple layers and involves non-linearity, activation only represents a special case. To take advantage of this, they propose to apply different channel-wise scales to weights, thus recovering accuracy.
    \item \textbf{Hessian:} \citet{NIPS1989_6c9882bb} first propose to use Hessian to calculate the change in loss:
    \begin{align}
    \label{eqn:Hessian}
        \Delta F\approx\frac{1}{2}(\Tilde{\wv}-\wv)^\top\Hv(\Tilde{\wv}-\wv),
    \end{align}
    since the gradient is close to zero when the model is well pre-trained. Following this line of research, later methods \cite{frantar2023gptqaccurateposttrainingquantization,pmlr-v235-kim24f,ding2024cbqcrossblockquantizationlarge} reduce the Hessian matrix to diagonal or block diagonal. Typically, to leverage Hessian requires to minimize Eq. (\ref{eqn:Hessian}), for example, by gradient descending \cite{tseng2024qtip} or K-means \cite{pmlr-v235-kim24f}.
\end{compactitem}

\section{Analysis of Existing Sensitivity Metrics}
The core of sensitivity is to convert element-wise quantization errors into estimation of $\Delta F$, thus analyzing the importance of each dimension. The sensitivity should accurately predict $\Delta F$, so we can specially choose the quantized weight $\Tilde \wv$ to minimize $\Delta F$. Unfortunately, our empirical study shows that Eq.(\ref{eqn:first-and-second}) is very inaccurate for LLMs.




\subsection{Precision of the approximation}
We run a simple experiment to quantize the 16-layer Llama 3.2 1B \cite{grattafiori2024llama3herdmodels} model with the simple codebook-based quantization method SqueezeLLM. We quantize the model to 4-bit with no detached outliers. We investigate two settings: (1) only quantize one layer, while leaving all other layers in full precision; (2) quantize all layers. For each setting, we compute the actual $\Delta F$ as well as the first order and second order term on the right hand side of Eq.~(\ref{eqn:first-and-second}).
Here, the Hessian matrix is computed by Fisher information approximation: $\Hv\approx -\frac{1}{n}\sum_i^{n} \nabla f_{\wv}(\wv,\xv_{[i]}) (\nabla f_{\wv}(\wv,\xv_{[i]}))^\top$.
The result is listed in \cref{table:first-order-and-second-order}. 
We have several observations, which may be surprisingly different from the common belief: 

\begin{table}[htb!]
\caption{First-order, second-order term and actual $\Delta F$ in \cref{eqn:first-and-second}.
}
\label{table:first-order-and-second-order}
\begin{center}
\begin{small}
\begin{tabular}{lccc}
\toprule
\makecell{\textbf{Quan-}\\\textbf{tized}\\\textbf{Layer}} & \textbf{First-order} & \textbf{Second-order} & \textbf{Actual} $\Delta F$ \\
\midrule
1                 & 7.10E-04        & -5.98E-06       & 6.88E-03  \\
2                 & -6.58E-05       & -4.54E-06       & 4.45E-03  \\
3                 & -3.21E-04       & -3.66E-06       & 3.67E-03  \\
4                 & -5.04E-04       & -3.68E-06       & 3.82E-03  \\
5                 & -7.00E-04       & -3.75E-06       & 3.72E-03  \\
6                 & -6.29E-04       & -3.61E-06       & 4.27E-03  \\
7                 & -2.04E-04       & -3.63E-06       & 5.06E-03  \\
8                 & 6.82E-05        & -3.60E-06       & 5.59E-03  \\
9                 & 5.75E-05        & -3.97E-06       & 6.85E-03  \\
10                & 2.86E-04        & -4.10E-06       & 7.78E-03  \\
11                & -6.43E-04       & -3.66E-06       & 6.57E-03  \\
12                & 8.29E-04        & -2.95E-06       & 6.81E-03  \\
13                & 6.14E-04        & -2.80E-06       & 5.83E-03  \\
14                & 1.30E-03        & -2.65E-06       & 6.57E-03  \\
15                & -2.52E-04       & -2.84E-06       & 5.30E-03  \\
16                & 3.47E-04        & -5.05E-06       & 9.79E-03  \\
All               & 8.92E-04        & -6.05E-05       & 1.00E-01  \\
\bottomrule
\end{tabular}
\end{small}
\end{center}
\end{table}

\paragraph{Observation 1: Second-order approximation is not accurate}
The most important observation is that the second-order approximation Eq.~(\ref{eqn:first-and-second}) deviates significantly from the actual change of loss $\Delta F$. Note that $\Delta F$ is always positive, which aligns with the intuition that the loss should increase after quantization. However, the sum of first- and second- order terms is much smaller than $\Delta F$, by more than an order of magnitude, implying that \emph{the second-order approximation significantly underestimates the change of loss}. The sum is even negative for sum layers, meaning that second-order approximation incorrectly predicts the decrease in loss after quantization. 

A direct consequence is that, neither the gradient-based approximation Eq.~(\ref{eqn:gradiet}) nor the Hessian-based approximation Eq.~(\ref{eqn:Hessian}) is accurate. Moreover, as the first- and second- order terms are both nonzero, both of them cannot be omitted. 

\paragraph{Observation 2: Layers are interdependent}
Another observation is that, the $\Delta F$ for quantizing each layer separately does not sum to the $\Delta F$ for quantizing all layers at once. While the first-order terms should be summable, all the higher-order terms can reflect correlations between layers. For example, the non-diagonal elements in the Hessian. This implies that separately computing each layer's sensitivity may not work.

\subsection{Convergence Radius}
The reason behind these counterintuitive observations is the convergence radius of Taylor's expansion. Eq.~(\ref{eqn:first-and-second}) requires the LLM to exhibit ideal mathematical properties: the existence of Hessian, and closeness between $\Tilde{\wv}$ and $\wv$. The definition of Hessian can be extended to non-differentiable activation functions such as ReLU~\cite{li2017convergence}. However, the closeness between $\Tilde{\wv}$ and $\wv$ is not guaranteed. Specifically, the Taylor series is meaningful only when $\tilde{\mathbf{w}}$ falls within the convergence radius of $\mathbf{w}$. However, LLMs have billions of parameters, leading to substantial total quantization errors and large $\norm{\Tilde{\wv}-\wv}$ distance.

To verify this, we gradually move $\Tilde{\wv}$ towards $\wv$ by an interpolation $\wv'=(1-\lambda)\wv+\lambda\Tilde{\wv}$. When $\lambda\rightarrow 0$, we can narrow the distance $\norm{\wv'-\wv}\rightarrow 0$. We list $\Delta F=F(\wv')-F(\wv)$ with different $\lambda$ in \cref{table:lambda}. Results show as $\lambda$ decreases, the second-order approximation becomes more accurate, and the Hessian term diminishes. 
When $\lambda\le 10^{-2}$, the gradient term alone is quite accurate. 

\begin{table}[htb!]
\caption{Actual $\Delta F$ with different $\lambda$.}
\label{table:lambda}
\begin{center}
\begin{small}
\begin{tabular}{lccc}
\toprule
$\lambda$ & \textbf{First-order} & \textbf{Second-order} & \textbf{Actual} $\Delta F$ \\
\midrule
1E-1                 & 8.92E-5 & -6.05E-7 & 1.00E-3 \\
5E-2                 & 4.46E-5 & -1.51E-7 & 2.73E-4 \\
1E-2                 & 8.92E-6 & -6.05E-9 & 1.81E-5 \\
5E-3                 & 4.46E-6 & -1.51E-9 & 6.68E-6 \\
1E-3                 & 8.92E-7 & -6.05E-11 & 9.54E-7 \\
\bottomrule
\end{tabular}
\end{small}
\end{center}
\end{table}

\section{Post-quantization Integral: An Accurate Sensitivity Metric}
\label{sec:pqi}

In this section, we propose a novel sensitivity metric, Post-quantization Integral (PQI), which can \emph{accurately} predict the change in loss in a \emph{fine-grained} way of quantizing each dimension. Most importantly, PQI works by decomposing the distant path from $\wv$ to $\Tilde{\wv}$ into many small fragments, so each one can be approximated accurately by Taylor formula. Besides, both $\wv$ and $\Tilde{\wv}$ are considered, so sensitivity is more precise.
Formally, given that $\Delta F$ can be rewritten as the integral
\begin{align}
\label{eqn:integral}
    \Delta F&=F(\Tilde{\wv})-F(\wv)=\int_{C}(\nabla_{\tv} F(\tv))^\top d\tv\nonumber
    \\
    &=\left[\int_{0}^{1}\nabla F((1-t)\wv+t\Tilde{\wv}) dt \right]^\top(\Tilde{\wv}-\wv),
\end{align}
where $C$ be any trajectory\footnote{Since $\nabla F$ is a potential function, the result is actually not related to $C$.} from $\wv$ to $\Tilde{\wv}$. we define PQI as
\begin{align}
\label{eqn:pqi}
    \vv_{PQI}=\abs{\int_{0}^{1}\nabla F((1-t)\wv+t\Tilde{\wv}) dt},
\end{align}
and we define:
\begin{align}
\label{eqn:pqi-loss}
    \Delta F_{PQI}=\vv_{PQI}^\top\abs{\Tilde{\wv}-\wv}.
\end{align}
Here, $\abs{\cdot}$ means element-wise absolute value.
Note that in $\vv_{PQI}$ and $\Delta F_{PQI}$, we omit the sign of each dimension, only focusing on its magnitude. Thus, $\Delta F_{PQI}$ stands for an approximate upper bound of $\Delta F$. This is because when counting total $\Delta F$, positive and negative entries in Eq. (\ref{eqn:integral}) will offset each other. However, this may lead to overfitting problems. To solve this, we need to take both positive and negative elements as an equal impact on the quantization results, and maximize generalization performance. Since we do not differentiate the sign when evaluating sensitivity, $\Delta F_{PQI}$ is slightly different from the original definition.


In practice, We complete the numerical integration with the following rectangle approximation:
\begin{align}
\label{eqn:pqi}
    \vv_{PQI}=\frac{1}{N}\sum_{i=1}^{N} \abs{\nabla F\left(\wv+\frac{i}{N}(\Tilde{\wv}-\wv)\right)},
\end{align}
where $N$ is the number of intervals. \cref{table:pqi} shows the taking $N\ge32$ can control the error within $\sim$0.1\%.

PQI can be used as a fine-grained metric for predicting the effect of quantizing \emph{each dimension} to the loss. Quantizing an element $w_i$ to $\tilde w_i$ yields $F$ to increase by $v_{PQI, i} \abs{\tilde w_i - \tilde w_i}$. Moreover, notice that the change in loss Eq.~(\ref{eqn:pqi-loss}) is \emph{linear} with $\tilde \wv$. Therefore, the effect of quantizing multiple dimensions is directly summable. Hence we can leverage PQI to estimate the effect of quantizing a group, a channel, a layer, or a block of weights by summing up $v_{PQI, i} \abs{\tilde w_i - \tilde w_i}$ for each element $w_i$ in the corresponding group.

\paragraph{Numerical Experiments}
We repeat the experiment of Llama 3.2 1B model but use our proposed PQI and counting average $\Delta F_{PQI}$ for each layer. We choose $N=32$ and sample a batch of 1000 sentences from WikiText-2 \cite{merity2017pointer}, while we keep the rest settings the same as \cref{table:first-order-and-second-order}. Results in \cref{table:layer-pqi} show that taking 32 intervals for integral yields perfect accuracy. Furthermore, we observe: 1) shallow layers have larger average $\Delta F_{PQI}$, while deeper layers typically have smaller $\Delta F_{PQI}$; 2) in a single layer, different sublayers exhibit different sensitivity. Specifically, \texttt{v\_proj} has the largest average $\Delta F_{PQI}$, denoting greatest impact on final results, followed by \texttt{k\_proj} and \texttt{o\_proj}.



\begin{table}[htb!]
\caption{Predicted $\Delta F$ with intervals we split. For reference, the actual $\Delta F(\wv)$ on this dataset should be 0.1024.}
\label{table:pqi}
\begin{center}
\begin{small}
\begin{tabular}{lcc}
\toprule
\textbf{Intervals} & \textbf{Predicted} $\Delta F$ & \textbf{Error} \\
\midrule
4                 & 1.042E-1 & 1.72E-2 \\
8                 & 1.032E-1 &  8.39E-4\\
16                 & 1.028E-1 & 3.90E-4 \\
32                 & 1.026E-1 & 1.62E-4 \\

\bottomrule
\end{tabular}
\end{small}
\end{center}
\end{table}

The only consideration about PQI now is that it cannot be used to make predictions ahead of quantization, since $\tilde \wv$ is required in advance to construct the path from $\wv$ to $\tilde \wv$. Therefore, we cannot leverage PQI directly to determine the optimal $\tilde \wv$ that minimizes the loss increment. Our solution is to first obtain a ``draft'' version of the quantized model with a traditional sensitivity metric, and then refine the quantization with PQI, which we shall discuss next. 

\begin{table*}[htb!]
\caption{Element-wise average $\Delta F_{PQI}$ of different layers and sublayers.}
\label{table:layer-pqi}
\begin{center}
\begin{small}
\begin{tabular}{lccccccc}
\toprule
\textbf{Layer} & \textbf{Q}      & \textbf{K}      & \textbf{V}      & \textbf{O}      & \textbf{Gate}   & \textbf{Up}     & \textbf{Down}   \\
\midrule
1     & 4.53E-08 & 9.93E-08 & 1.59E-07 & 9.13E-08 & 4.22E-08 & 4.99E-08 & 5.31E-08 \\
5     & 4.16E-08 & 6.66E-08 & 1.07E-07 & 7.37E-08 & 2.57E-08 & 4.14E-08 & 4.37E-08 \\
8     & 3.83E-08 & 6.11E-08 & 9.94E-08 & 8.83E-08 & 2.46E-08 & 4.01E-08 & 4.72E-08 \\
11    & 2.63E-08 & 4.53E-08 & 7.88E-08 & 4.67E-08 & 2.90E-08 & 3.78E-08 & 4.61E-08 \\
\bottomrule
\end{tabular}
\end{small}
\end{center}
\end{table*}

\section{ReQuant: A Sensitivity-aware Quantization Pipeline}
\label{sec:method}
In this section, we take advantage of PQI by proposing ReQuant, a quantization pipeline orthogonal to most of previous LLM quantization methods, and can be easily combined with the original quantization process. The core of this pipeline is a Dense-and-Sparse detach:
\begin{align}
    \Tilde{\wv} = Q(\wv-\wv_o,\vv)+\wv_o+\wv_s,
\end{align}
where $Q(\wv-\wv_o,\vv)$ is a low-precision dense weight vector, while $\wv_o$ and $\wv_s$ are high-precision sparse outliers and significant weights, respectively. We start by explaining the two sparse components in our pipeline. Then, we present an overview of the whole workflow.

\subsection{Self-adaptive Outlier Selection}
\label{sec:outlier-selection}
One of the biggest challenges in LLM quantization is outliers, and previous studies have proved that removing outliers helps to promote the model's performance \cite{MLSYS2024_42a452cb,pmlr-v235-kim24f,shao2024omniquant}. 
However, the biggest problem we identify from previous studies is that they use the same ratio or clamping threshold for all layers in the model. Recall that in \cref{sec:pqi}, we have shown that some specific layers have larger average $\Delta F_{PQI}$, indicating a larger impact on the final accuracy loss. This leads to the assumption that layers with larger $\Delta F_{PQI}$ should be allocated with more bits to quantize, correspondingly higher outlier ratio in our Dense-and-Sparse decomposition. Thus, utilizing the same ratio or threshold results in suboptimal final results. To strike a balance among different layers, we devise a ratio searching process based on PQI.
The key to this method is to dynamically adjust each layer's outlier ratio according to layer-wise $\Delta F_{PQI}$.
To tune the optimal proportion, we add a temperature factor $t$ to the fractions, and run a grid search on it:
\begin{align}
\label{eqn:outlier-search}
    outlier\_num^{(i)}=\operatorname{dim}(\wv)\times r_o\% \times \frac{(\Delta F_{PQI}^{(i)})^{t}}{\sum_{i} (\Delta F_{PQI}^{(i)})^{t}},
\end{align}
where $r_o\%=\dim(\wv_o)/\dim(\wv)$ is the global outlier ratio. Superscripts ${(i)}$ indicate parameters of the $i$-th layers. The relevant process is formulated in Algorithm \ref{alg:o-search}.

\begin{algorithm}[htb!]
   \caption{Outlier Ratio Search}
   \label{alg:o-search}
\begin{algorithmic}
   \STATE {\bfseries Input:} layers $l$, global outlier ratio $r_o\%$, weight vector $\wv$, original sensitivity $\vv$, search step $\alpha$, pre-quantized $\Tilde{\wv}$
   \STATE $t_{best}=0,loss_{min}=\infty$
   \STATE Calculate $\{\Delta F_{PQI}^{(i)}\}_{1\leq i \leq l}$ via \cref{eqn:pqi,eqn:pqi-loss}
   \FOR{$t=0;t<1;t=t+\alpha$}
   
   \FOR{$i=1$ {\bfseries to} $l$}
   \STATE Calculate $outlier\_num^{(i)}$ via \cref{eqn:outlier-search}
   \STATE Select the largest $outlier\_num^{(i)}$ elements from $\wv^{(i)}$ as $\wv_o^{(i)}$
   \STATE $\Tilde{\wv}=Q(\wv-\wv_o,\vv)+\wv_o$
   \IF{$F(\Tilde{\wv})<loss_{min}$}
   \STATE $t_{best}=t,loss_{min}=F(\Tilde{\wv})$
   \ENDIF
   \ENDFOR   
   \ENDFOR
   \STATE Calculate $\{outlier\_num^{(i)}\}_{1\leq i \leq l}$ via \cref{eqn:outlier-search}
   \STATE \textbf{return} $\{outlier\_num^{(i)}\}_{1\leq i \leq l}$
\end{algorithmic}
\end{algorithm}

\subsection{Step-wise Significant Weights Detach}
\label{sec:significant-detach}
Though outlier selection improves performance, it may not be enough only to clip outliers from weights. Intuitively, the element-wise importance of weights is implied by an element-wise product: $\vv_{PQI}\odot\abs{\Tilde{\wv}-\wv}$.
As shown in \cref{table:distribution}, some ``significant'' weights are crucial to the final output, with top 5.25\% significant weights covering over 30\% of total $\Delta F_{PQI}$. To further improve performance, we devise a \emph{greedy search} algorithm (\cref{alg:s-search}) to gradually detach $r_s\%$ important weights via PQI. $\beta \% \leq r_s\%$ is a small fraction we detach in a single pass. Here, $r_s\%$ should be an integer multiple of $\beta\%$. The reason why we choose to ``gradually'' detach is to pick out the most important weights for current $\Tilde{\wv}$, as well as minimize the estimation error. In practice, we notice that in most circumstances, even if we pick them out all at once (\ie, $\beta=r_s$), the final performance would not degrade much.

\begin{table}[h]
\caption{The proportion of significant weights we choose and how much they can cover in total $\Delta F_{PQI}$.}
\label{table:distribution}
\begin{center}
\begin{tabular}{cc}
\toprule
\makecell{\textbf{Proportion of} \\ \textbf{Significant Weights}} & \makecell{$\Delta F_{PQI}$ \textbf{percentage} } \\
\midrule
0.15\% & 4.53\% \\
0.71\% & 11.29\% \\
5.25\% & 34.06\% \\
\bottomrule
\end{tabular}
\end{center}
\end{table}

\begin{algorithm}[htb!]
   \caption{Significant Weight Search}
   \label{alg:s-search}
\begin{algorithmic}
   \STATE {\bfseries Input:} global significant weight ratio $r_s\%$, weight vector $\wv$, Re-quantized weights $\Tilde{\wv}$, search step $\beta\%$
   \STATE $\wv_s=\mathbf{0}$
   \FOR{$i=0;i<r_s/\beta;i=i+1$}
   \STATE Calculate $\vv_{PQI}$ via \cref{eqn:pqi}
   \STATE Calculate element-wise $\Delta F_{PQI}$ by $\vv_{PQI}\odot\abs{\Tilde{\wv}-\wv}$
   \STATE Select the top $\beta\%$ elements with the largest $\Delta F_{PQI}$ from $(\wv-\Tilde{\wv})$ as $\wv'_s$
   \STATE $\Tilde{\wv}=\Tilde{\wv}+\wv'_s,\wv_s=\wv_s+\wv'_s$
   \ENDFOR
   \STATE \textbf{return} $\wv_s$
\end{algorithmic}
\end{algorithm}

\subsection{Quantization Pipeline}
As a combination and conclusion of \cref{sec:outlier-selection,sec:significant-detach}, we now present an overview of our proposed workflow:

\begin{compactitem}[$\bullet$]
    \item[1)] Pre-quantize $\Tilde{\wv} = Q(\wv,\vv)$;
    \item[2)] Run \cref{alg:o-search} to get $\{outlier\_num^{(i)}\}_{1\leq i \leq l}$;
    \item[3)] Select outliers $\{\wv_o^{(i)}\}_{1\leq i \leq l}$ from $\{\wv^{(i)}\}_{1\leq i \leq l}$;
    \item[4)] Re-quantize $\Tilde{\wv} = Q(\wv-\wv_o,\vv)+\wv_o$;
    \item[5)] Recover significant weights $\wv_s$ with \cref{alg:s-search};
    \item[6)] Complete quantization: $\Tilde{\wv} = Q(\wv-\wv_o,\vv)+\wv_o+\wv_s$.
\end{compactitem}

\begin{figure}[htb!]
    \centering
    \includegraphics[width=\linewidth]{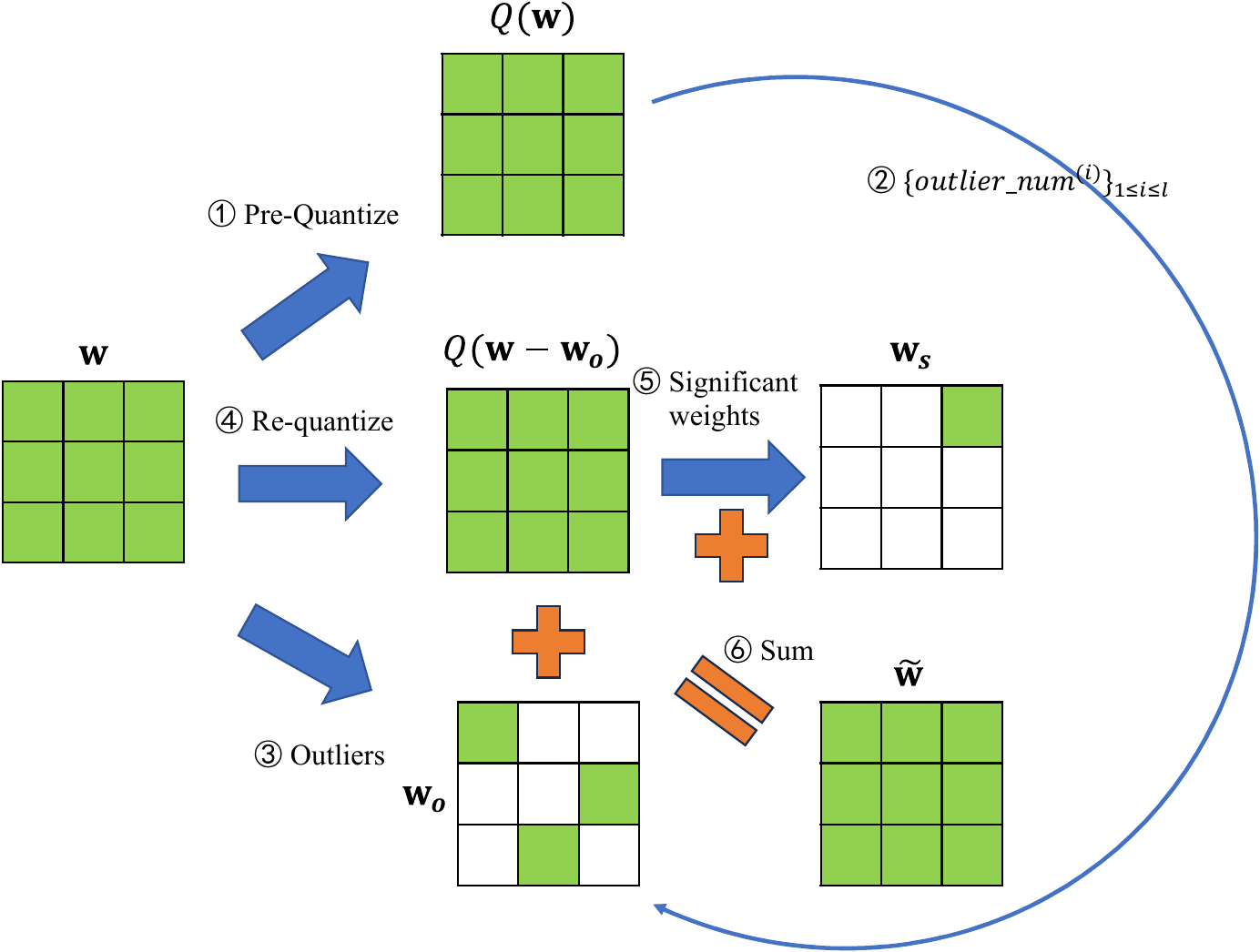}
    \caption{ReQuant pipeline.}
    \label{fig:workflow}
\end{figure}

It is worth noticing that while $\wv_o$ detach the outliers of $\wv$, $\wv_s$ actually detaches the outliers of $\vv_{PQI}\odot\abs{\Tilde{\wv}-\wv}$. Besides, the actual quantization happens in step 4. Step 1 is merely a pre-quantization step to determine the outlier ratio $r_o$. $\wv_s$ is acquired after quantization, aiming to ``recover'' the accuracy of important weights. The scheme of our proposed workflow is illustrated in \cref{fig:workflow}.

\begin{table*}[htb!]
\caption{QTIP results for Llama 3.2 1B Base/Instruct models. The entries share the same meaning as \cref{table:math}.}
\label{table:qtip}
\vspace{-6mm}
\centering{
\footnotesize{
    \vspace{2mm}
    
    \begin{subtable}
        \centering

        \vspace{-1mm}
        \begin{tabular}{c|c|c|cc|c|c|c|c}
         \toprule
            \multirow{2}{*}{\textbf{Precision}} & \multirow{2}{*}{\textbf{Method}} & \multirow{2}{*}{\textbf{Calib Set}} & \multicolumn{2}{c|}{\textbf{Sparsity}} & \multirow{2}{*}{\textbf{Bits}}  & {\textbf{Mem}} & {\textbf{Base}} & {\textbf{Instruct}} \\
            & & & $r_o$ & $r_s$ & & (GB) & Wiki2$\downarrow$ & MATH$\uparrow$\\
            \midrule
            \midrule
            full & Baseline & - & - & - & 16 & 2.30 & 9.75 & 29.30\\
            \midrule
            \multirow{2}{*}{2-bit}& QTIP & RedPajama & - & - & 2.02 & 1.40 & 18.67 & 0.78 \\
            & QTIP+ReQuant & RedPajama/WikiText-2/Tulu 3 & 0 & 0.5 & 2.26 & 1.47 & \textbf{16.01} & \textbf{2.68} \\
            \midrule
            \multirow{2}{*}{3-bit}& QTIP & RedPajama- & - & - & 3.02 & 1.72 & 11.17 & 18.78 \\
            & QTIP+ReQuant & RedPajama/WikiText-2/Tulu 3 & 0 & 0.5 & 3.26 & 1.80 & \textbf{10.83} & \textbf{20.06}\\
            \midrule
            \multirow{2}{*}{4-bit} & QTIP & RedPajama & - & - & 4.02 & 2.05 & 10.12 & 26.38 \\
            & QTIP+ReQuant & RedPajama/WikiText-2/Tulu 3 & 0 & 0.5 & 4.26 & 2.13 & \textbf{10.06} & \textbf{27.36} \\
        \bottomrule
        
        \end{tabular}

    \end{subtable}
         
     }
     }
\end{table*}

\section{Experiment}
In this section, we apply ReQuant on Llama 3.2 1B and 3B small models. We pick out three representative grouping-based and codebook-based quantization methods: AWQ, SqueezeLLM, and QTIP, and evaluate ReQuant on them.

\subsection{Accuracy Results}
For AWQ, following its original settings, we sample 100 sequences of 2048 length from Pile \cite{gao2020pile800gbdatasetdiverse} dataset as its calibration set and for the ReQuant process. To reach comparable total bits, we choose 0.25\% sparse ratio with group size of 256, while the baseline adopts 128 group size. Note that both increasing the group size and increasing sparsity improves the results, and primary results show that 256 grouping with 0.25\% sparsity gives optimal results. For SqueezeLLM, we use sampled 100 sentences from WikiText-2 with 2048 sequence length as calibration set to calculate Hessian matrix and PQI. Since SqueezeLLM adopts the same Dense-and-Sparse decomposition format as ReQuant, we simply replace the counterpart with our detached weights. For QTIP, we use RedPajama \cite{weber2024redpajama} for its original calibration. We use Tulu 3 \cite{lambert2024tulu3} dataset for supervised fine-tuning models to calculate PQI, and use WikiText-2 dataset for base models.

\paragraph{Perplexity Results for Pre-trained Models} For Llama 3.2 1B and 3B models, we evaluate the model on WikiText-2 test set with sequence length 2048. Results are presented in \cref{table:ppl,table:qtip}. Results show significant improvement of ReQuant, highlighting a 1.38 decrease on the 3-bit 1B model with AWQ and a 0.56 decrease with SqueezeLLM. For 3B model, the improvement is minor since larger model are easier to quantize for the baselines.

\paragraph{Few-shot Results for Instruction Following Models}

For instruction following models, we evaluate on the 4-shot MATH generation task. 
Notably, AWQ+ReQuant and SqueezeLLM+ReQuant show significant improvements: AWQ+ReQuant in 4-bit quantization achieves a MATH score of 24.32, while SqueezeLLM+ReQuant achieves a higher score of 24.74, demonstrating superior performance. Besides, ReQuant helps to restore severe loss in the model's accuracy: SqueezeLLM baseline, which fails to present the desired answer format, can yield feasible outputs after applying ReQuant. Results are shown in \cref{table:math,table:qtip}.
\begin{table*}[htb!]
\caption{WikiText-2 perplexity for base models and 4-shot MATH evaluation for instruction following models. ``Fail'' means failure to parse model's output due to garbled characters.}
\label{table:ppl}
\label{table:math}
\vspace{-6mm}
\centering{
\footnotesize{
\setlength{\tabcolsep}{7pt}{
    \vspace{2mm}
    
    \begin{subtable}
        \centering
        \scriptsize{

        \vspace{-1mm}
        \begin{tabular}{c|c|cc|c|c|c|c|c|c|c|c}
         \toprule
            {\textbf{Llama 3.2 1B Base/Instruct}} & \multicolumn{3}{c|}{\textbf{Hyperparameters}} &\multicolumn{4}{c|}{\textbf{3-bit}} & \multicolumn{4}{c}{\textbf{4-bit}} \\
            \midrule
            \multirow{2}{*}{\textbf{Method}} & \multirow{2}{*}{\textbf{Calib Set}} & \multicolumn{2}{c|}{\textbf{Sparsity}} & \multirow{2}{*}{\textbf{Bits}}  & {\textbf{Mem}} & {\textbf{Base}} & {\textbf{Instruct}} & \multirow{2}{*}{\textbf{Bits}} & {\textbf{Mem}} & {\textbf{Base}} & {\textbf{Instruct}} \\
            & & $r_o$ & $r_s$ & & (GB) & Wiki2$\downarrow$ & MATH$\uparrow$ & & (GB) & Wiki2$\downarrow$ & MATH$\uparrow$ \\
            \midrule
            \midrule
            Baseline & - & - & - & 16 & 2.30 & 9.75 & 29.30 & 16 & 2.30 & 9.75 & 29.30 \\
            \midrule
            AWQ (g128) & Pile & - & - & 3.25 & 0.86 & 16.74 & fail & 4.25 & 0.97 & 10.84 & 22.82\\
            AWQ (g256)+ReQuant & Pile & 0.25 & 0 & 3.25 & 0.86 & \textbf{15.36} & fail & 4.25 & 0.97 & \textbf{10.65} & \textbf{24.32}\\
           \midrule
            SqueezeLLM & WikiText-2 & 0.45 & 0.05 & 3.25 & 0.86 & 13.86 & 11.28 & 4.25 & 0.97 & 10.51 & fail\\
            SqueezeLLM+ReQuant & WikiText-2 & 0.45 & 0.05 & 3.25 & 0.86 & \textbf{13.30} & \textbf{14.18} & 4.25 & 0.97 & \textbf{10.43} & \textbf{24.74}\\

        \bottomrule
        
        \end{tabular}
        }

    \end{subtable}
    \begin{subtable}
        \centering
        \scriptsize{
        \vspace{-1mm}
        \begin{tabular}{c|c|cc|c|c|c|c|c|c|c|c}
         \toprule
            {\textbf{Llama 3.2 3B Base/Instruct}} & \multicolumn{3}{c|}{\textbf{Hyperparameters}} &\multicolumn{4}{c|}{\textbf{3-bit}} & \multicolumn{4}{c}{\textbf{4-bit}} \\
            \midrule
            \multirow{2}{*}{\textbf{Method}} & \multirow{2}{*}{\textbf{Calib Set}} & \multicolumn{2}{c|}{\textbf{Sparsity}} & \multirow{2}{*}{\textbf{Bits}}  & {\textbf{Mem}} & {\textbf{Base}} & {\textbf{Instruct}} & \multirow{2}{*}{\textbf{Bits}} & {\textbf{Mem}} & {\textbf{Base}} & {\textbf{Instruct}} \\
            & & $r_o$ & $r_s$ & & (GB) & Wiki2$\downarrow$ & MATH$\uparrow$ & & (GB) & Wiki2$\downarrow$ & MATH$\uparrow$ \\
           \midrule
           \midrule
           Baseline & - & - & - & 16 & 5.98 & 7.81 & 44.92 & 16 & 5.98 & 7.81 & 44.92\\
           \midrule
            AWQ (g128) & Pile & - & - & 3.25 & 1.80 & 10.30 & 29.64 & 4.25 & 2.13 & 8.22 & \textbf{42.88}\\
            AWQ (g256)+ReQuant & Pile & 0.25 & 0 & 3.24 & 1.80 & \textbf{9.98} & \textbf{35.08} & 4.24 & 2.13 & \textbf{8.20} & 42.20\\
           \midrule
            SqueezeLLM & WikiText-2 & 0.45 & 0.05 & 3.24 & 1.80 & \textbf{9.39} & 33.80 & 4.24 & 2.13 & \textbf{8.12} & \textbf{43.06}\\
            SqueezeLLM+ReQuant & WikiText-2 & 0.45 & 0.05 & 3.24 & 1.80 & 9.47 & \textbf{35.34} & 4.24 & 2.13 & 8.14 & 42.24 \\

        \bottomrule
        
        \end{tabular}
        
     \vspace{-1mm}
        }
    
     \end{subtable}

     }
     }
     }
\end{table*}

\subsection{Ablation Study}
In this part, we explore the effectiveness of $\wv_o$ and $\wv_s$ separately. We evaluate our method with Llama 3.2 1B base model on WikiText-2 train and test set. Results are shown in \cref{table:ablation}.
\paragraph{Outlier Selection}
To validate the effect of outlier selection, we compare 0.45\% outlier results with the counterpart in SqueezeLLM. Results show that the uneven outlier ratio among different layers improves test perplexity from 10.62 to 10.52, while train perplexity from 11.15 to 11.02. Considering the model size and precision, we believe this improvement is significant. We also add a ``rand'' baseline which randomly picks $\wv_o$ and $\wv_s$. Comparison shows that ReQuant has the best performance to identify outliers.

\paragraph{Significant Weight Detach}
The success of significant weight detach rely on accurately identifying important weights. To examine the influence of searching steps $\frac{r_s}{\beta}$ on final results, we list results when $r_s/\beta=1,2,4$ and compare train and test perplexity. Results show that increasing greedy searching steps helps to improve performance, with $r_s/\beta=2$ improves test perplexity by 0.03. However, doubling 
searching steps also doubles the integral computation cost. Since more steps only provides minimum improvement, we choose $r_s/\beta=2$.

\begin{table}[htb!]
\caption{Ablation results on WikiText-2 perplexity. The ``rand'' line indicates that $\wv_o$ and $\wv_s$ are picked out randomly from the weights.}
\label{table:ablation}
\begin{center}
\begin{tabular}{cclcc}
\toprule
$r_o$ & $r_s$ & \textbf{Comment} & \textbf{Train PPL} & \textbf{Test PPL} \\
\midrule
- & - & bfloat16 & 10.20 & 9.75 \\
0.45 & 0.05 & & 10.95 & 10.45 \\
\midrule
0.45 & 0 & & 11.02 & 10.52 \\
0.45 & 0 & SqueezeLLM & 11.15 & 10.62 \\
0 & 0.05 & & 11.15 & 10.65 \\
0 & 0 & & 11.30 & 10.80 \\
\midrule
0.45 & 0.05 & rand & 11.28 & 10.77 \\
\midrule
0.45 & 0.05 & $\beta=0.0125$ & 10.94 & 10.42 \\
0.45 & 0.05 & $\beta=0.025$ & 10.94 & 10.42 \\
0.45 & 0.05 & $\beta=0.05$ & 10.95 & 10.45 \\
\bottomrule
\end{tabular}
\end{center}
\end{table}

\subsection{Inference Speed}
The inference speed of Dense-and-Sparse decomposition is tested with Llama 3.2 1B and 3B instruction following models. For sparse matrix multiplication, we use cuSPARSE\footnote{\url{https://docs.nvidia.com/cuda/cusparse}}, while for the dense quantized weight multiplication, we keep the methods' original inference kernels. We report the latency of a single linear layer and for the the whole model. We use the same settings for all experiments: an input sentence of 128 length, and 2048 generated tokens. For all experiments, we use RTX 3090 GPUs. Our Dense-and-Sparse decomposition meets the baseline inference speed of its equivalents \cite{pmlr-v235-kim24f} and is slightly (1.2$\sim$1.4x) slower than other inference frameworks, which is reasonable compared with similar studies \cite{li2023losparsestructuredcompressionlarge,pmlr-v235-kim24f}; see \cref{table:inference}.

\begin{table}[htb!]
\caption{Inference speed of Dense-and-Sparse decomposition.}
\label{table:inference}
\begin{center}
\resizebox{0.5\textwidth}{!}{
\begin{tabular}{lllccc}
\toprule


\multirow{2}{*}{\textbf{Model}} & \multirow{2}{*}{\textbf{Precision}} & \multirow{2}{*}{\textbf{Method}} & \textbf{Prefilling} & \textbf{Decoding} & \textbf{Total} \\
& & & (ms) & (ms) & (ms) \\
\midrule
1B & 4-bit & AWQ & \textbf{13} & \textbf{23768} & \textbf{23781} \\
1B & 4-bit & AWQ+PQI & 18 & 35204 &  35222 \\
\midrule
1B & 4-bit & SqueezeLLM & 86 & 47151 & 47237 \\
1B & 4-bit & SqueezeLLM+ReQuant & \textbf{29} & \textbf{45266} & \textbf{45295}\\
\midrule
1B & 3-bit & SqueezeLLM & \textbf{85} & 33657 & 33742\\
1B & 3-bit & SqueezeLLM+ReQuant & 86 & \textbf{32568} & \textbf{32654} \\
\midrule
3B & 4-bit & AWQ & 31 & 59631 & 59662 \\
3B & 4-bit & AWQ+ReQuant & 31 & \textbf{59174} & \textbf{59205} \\
\midrule
3B & 4-bit & SqueezeLLM & 230 & 56882 & 57112 \\
3B & 4-bit & SqueezeLLM+ReQuant & \textbf{68} & \textbf{56372} & \textbf{56440} \\
\midrule
3B & 3-bit & SqueezeLLM & 229 & 56343 & 56572\\
3B & 3-bit & SqueezeLLM+ReQuant & 229 & \textbf{54640} & \textbf{54869}\\

\bottomrule
\end{tabular}
}
\end{center}
\end{table}
\section{Discussion and Conclusion}
In this study, we discuss the accuracy of sensitivity metrics in PTQ. We reveal the inaccuracy of the classic second order approximation of $\Delta F$ and point out the reason to be small convergence radius. To tackle this problem, we propose PQI as a resolution, which calculates the posterior element-wise quantization impact. To leverage PQI, we propose ReQuant, a workflow based on Dense-and-Sparse decomposition to boost the performance of a low-precision model. Our method is validated on popular AWQ, SqueezeLLM, and QTIP, with frontier Llama 3.2 small language models.

\paragraph{Future Work} Our proposed workflow mainly targets on recovering the important weights after quantization. While one can do this process iteratively to detach the desired significant elements, it still requires to store a sparse matrix, which consists of undesired bits, like the row and column indices. Meanwhile, the sparse matrix multiplication is a bottleneck in inference. Thus, combining the PQI analysis into the quantization process is left as future work. For example, one can iteratively compute PQI and tune the weight distribution according to it, thus getting rid of the sparse weights.


\section*{Impact Statement}
This paper presents work whose goal is to advance the field of Machine Learning. There are many potential societal consequences of our work, none which we feel must be specifically highlighted here.

\bibliography{example_paper}
\bibliographystyle{icml2025}

\newpage
\appendix
\onecolumn

\section{Limitations}
\label{limitations}
In practice, we identify two limitations. First, the Dense-and-Sparse decomposition requires the sparse matrix to store its weights and index. While the weights require at least 16 bits per element, the row and column indices require at least 16 bits for each, which is non-negligible. Due to this reason, we only adopt minimum sparsity (0.5\%). Besides, sparse matrix multiplication is slow, and may harm decoding speed during inference.

\section{Hyperparameters}

\begin{table}[htb!]
\caption{Experimental hyperparameters.}
\label{table:distribution}
\begin{center}
\begin{tabular}{llc}
\toprule
\textbf{Setting} & \textbf{Hyperparameter} & \textbf{Value} \\
\midrule
\multirow{5}{*}{AWQ}              & calib set (all) & Pile \\
              & calib sequence length & 2048 \\
              & $N$ & 32 \\
              & $n$ & 100 \\
              & $r_s/\beta$ & 2 \\
\midrule
\multirow{5}{*}{SqueezeLLM}       & calib set (all) & WikiText-2 \\
              & calib sequence length & 2048 \\
              & $N$ & 32 \\
              & $n$ & 100 \\
              & $r_s/\beta$ & 2 \\
\midrule
\multirow{5}{*}{QTIP}             & calib set (Hessian) & RedPajama \\
             & calib set (ReQuant, base models) & WikiText-2 \\
             & calib set (ReQuant, instruction following models) & Tulu 3 \\
              & $N$ & 32 \\
              & $r_s/\beta$ & 2 \\

\bottomrule
\end{tabular}
\end{center}
\end{table}

\section{Experiments compute resources}
For reference, the estimated GPU hours of computing integral are listed Table \ref{table:resource}.
\begin{table}[htb!]
\caption{GPU Hours of doing integral on A100.}
\label{table:resource}
\begin{center}
\begin{tabular}{lllc}
\toprule
            & \textbf{Calib Set} & $n$   & \textbf{GPU Hours} \\
\midrule
Llama 3.2 1B  & WikiText & 100     & 0.5       \\
Llama 3.2 3B  & WikiText & 100     & 1.5       \\
Llama 3.2 1B  & Tulu 3 & 2048     & 1.5       \\
Llama 3.2 3B  & Tulu 3 & 2048     & 3       \\
\bottomrule
\end{tabular}
\end{center}
\vskip -0.1in
\end{table}


\end{document}

%% file: common.tex
\usepackage{color}
\usepackage{bm}
\usepackage{hyperref}
\usepackage{amsmath,amsfonts,amsthm}
\usepackage{blindtext}
\usepackage{listings}
\usepackage{algorithm, algorithmic}
\usepackage{booktabs}
\usepackage{multirow}
\usepackage{multicol}
\usepackage{caption}
\usepackage{enumitem} 

\newcommand{\ie}{\emph{i.e.}}

\newlist{myitems}{enumerate}{3}
\setlist[myitems, 1]
{label=\arabic{myitemsi}.,leftmargin=15pt,labelwidth=10pt,labelsep=5pt,
topsep=0pt,parsep=0pt,partopsep=0pt,noitemsep
}

\usepackage{ifthen}

\newif\ifdebug
\debugfalse

\newcommand{\vect}[1]{\boldsymbol{\mathbf{#1}}}

\newcommand{\tv}{\vect t}

\newcommand{\vv}{\vect v}
\newcommand{\wv}{\vect w}
\newcommand{\xv}{\vect x}

\newcommand{\Hv}{\vect H}

\newcommand{\Tv}{\vect T}

\newcommand{\Dc}{\mathcal D}

\newcommand{\norm}[1]{\left\lVert#1\right\rVert}
\newcommand{\abs}[1]{\left|#1\right|}

\makeatletter
\newtheorem*{rep@theorem}{\rep@title}
\newcommand{\newreptheorem}[2]{%
	\newenvironment{rep#1}[1]{%
		\def\rep@title{#2 \ref{##1}}%
		\begin{rep@theorem}}%
		{\end{rep@theorem}}}
\makeatother

%% file: example_paper.bbl
\begin{thebibliography}{45}
\providecommand{\natexlab}[1]{#1}
\providecommand{\url}[1]{\texttt{#1}}
\expandafter\ifx\csname urlstyle\endcsname\relax
  \providecommand{\doi}[1]{doi: #1}\else
  \providecommand{\doi}{doi: \begingroup \urlstyle{rm}\Url}\fi

\bibitem[Abdin et~al.(2024)Abdin, Aneja, Behl, Bubeck, Eldan, Gunasekar, Harrison, Hewett, Javaheripi, Kauffmann, Lee, Lee, Li, Liu, Mendes, Nguyen, Price, de~Rosa, Saarikivi, Salim, Shah, Wang, Ward, Wu, Yu, Zhang, and Zhang]{abdin2024phi4technicalreport}
Abdin, M., Aneja, J., Behl, H., Bubeck, S., Eldan, R., Gunasekar, S., Harrison, M., Hewett, R.~J., Javaheripi, M., Kauffmann, P., Lee, J.~R., Lee, Y.~T., Li, Y., Liu, W., Mendes, C. C.~T., Nguyen, A., Price, E., de~Rosa, G., Saarikivi, O., Salim, A., Shah, S., Wang, X., Ward, R., Wu, Y., Yu, D., Zhang, C., and Zhang, Y.
\newblock Phi-4 technical report, 2024.
\newblock URL \url{https://arxiv.org/abs/2412.08905}.

\bibitem[Ashkboos et~al.(2024)Ashkboos, Mohtashami, Croci, Li, Cameron, Jaggi, Alistarh, Hoefler, and Hensman]{ashkboos2024quarotoutlierfree4bitinference}
Ashkboos, S., Mohtashami, A., Croci, M.~L., Li, B., Cameron, P., Jaggi, M., Alistarh, D., Hoefler, T., and Hensman, J.
\newblock Quarot: Outlier-free 4-bit inference in rotated llms, 2024.
\newblock URL \url{https://arxiv.org/abs/2404.00456}.

\bibitem[Chee et~al.(2023)Chee, Cai, Kuleshov, and De~Sa]{NEURIPS2023_0df38cd1}
Chee, J., Cai, Y., Kuleshov, V., and De~Sa, C.~M.
\newblock Quip: 2-bit quantization of large language models with guarantees.
\newblock In Oh, A., Naumann, T., Globerson, A., Saenko, K., Hardt, M., and Levine, S. (eds.), \emph{Advances in Neural Information Processing Systems}, volume~36, pp.\  4396--4429. Curran Associates, Inc., 2023.
\newblock URL \url{https://proceedings.neurips.cc/paper_files/paper/2023/file/0df38cd13520747e1e64e5b123a78ef8-Paper-Conference.pdf}.

\bibitem[Dettmers et~al.(2022{\natexlab{a}})Dettmers, Lewis, Belkada, and Zettlemoyer]{NEURIPS2022_c3ba4962}
Dettmers, T., Lewis, M., Belkada, Y., and Zettlemoyer, L.
\newblock Gpt3.int8(): 8-bit matrix multiplication for transformers at scale.
\newblock In Koyejo, S., Mohamed, S., Agarwal, A., Belgrave, D., Cho, K., and Oh, A. (eds.), \emph{Advances in Neural Information Processing Systems}, volume~35, pp.\  30318--30332. Curran Associates, Inc., 2022{\natexlab{a}}.
\newblock URL \url{https://proceedings.neurips.cc/paper_files/paper/2022/file/c3ba4962c05c49636d4c6206a97e9c8a-Paper-Conference.pdf}.

\bibitem[Dettmers et~al.(2022{\natexlab{b}})Dettmers, Lewis, Shleifer, and Zettlemoyer]{dettmers2022bit}
Dettmers, T., Lewis, M., Shleifer, S., and Zettlemoyer, L.
\newblock 8-bit optimizers via block-wise quantization.
\newblock In \emph{International Conference on Learning Representations}, 2022{\natexlab{b}}.
\newblock URL \url{https://openreview.net/forum?id=shpkpVXzo3h}.

\bibitem[Dettmers et~al.(2023)Dettmers, Pagnoni, Holtzman, and Zettlemoyer]{NEURIPS2023_1feb8787}
Dettmers, T., Pagnoni, A., Holtzman, A., and Zettlemoyer, L.
\newblock Qlora: Efficient finetuning of quantized llms.
\newblock In Oh, A., Naumann, T., Globerson, A., Saenko, K., Hardt, M., and Levine, S. (eds.), \emph{Advances in Neural Information Processing Systems}, volume~36, pp.\  10088--10115. Curran Associates, Inc., 2023.
\newblock URL \url{https://proceedings.neurips.cc/paper_files/paper/2023/file/1feb87871436031bdc0f2beaa62a049b-Paper-Conference.pdf}.

\bibitem[Ding et~al.(2024)Ding, Liu, Tu, Zhang, Li, Hu, Chen, Tang, Xiong, Yin, and Wang]{ding2024cbqcrossblockquantizationlarge}
Ding, X., Liu, X., Tu, Z., Zhang, Y., Li, W., Hu, J., Chen, H., Tang, Y., Xiong, Z., Yin, B., and Wang, Y.
\newblock Cbq: Cross-block quantization for large language models, 2024.
\newblock URL \url{https://arxiv.org/abs/2312.07950}.

\bibitem[Frantar \& Alistarh(2022)Frantar and Alistarh]{NEURIPS2022_1caf09c9}
Frantar, E. and Alistarh, D.
\newblock Optimal brain compression: A framework for accurate post-training quantization and pruning.
\newblock In Koyejo, S., Mohamed, S., Agarwal, A., Belgrave, D., Cho, K., and Oh, A. (eds.), \emph{Advances in Neural Information Processing Systems}, volume~35, pp.\  4475--4488. Curran Associates, Inc., 2022.
\newblock URL \url{https://proceedings.neurips.cc/paper_files/paper/2022/file/1caf09c9f4e6b0150b06a07e77f2710c-Paper-Conference.pdf}.

\bibitem[Frantar et~al.(2023)Frantar, Ashkboos, Hoefler, and Alistarh]{frantar2023gptqaccurateposttrainingquantization}
Frantar, E., Ashkboos, S., Hoefler, T., and Alistarh, D.
\newblock Gptq: Accurate post-training quantization for generative pre-trained transformers, 2023.
\newblock URL \url{https://arxiv.org/abs/2210.17323}.

\bibitem[Gao et~al.(2020)Gao, Biderman, Black, Golding, Hoppe, Foster, Phang, He, Thite, Nabeshima, Presser, and Leahy]{gao2020pile800gbdatasetdiverse}
Gao, L., Biderman, S., Black, S., Golding, L., Hoppe, T., Foster, C., Phang, J., He, H., Thite, A., Nabeshima, N., Presser, S., and Leahy, C.
\newblock The pile: An 800gb dataset of diverse text for language modeling, 2020.
\newblock URL \url{https://arxiv.org/abs/2101.00027}.

\bibitem[Grattafiori et~al.(2024)Grattafiori, Dubey, Jauhri, Pandey, Kadian, Al-Dahle, Letman, Mathur, Schelten, Vaughan, Yang, Fan, Goyal, Hartshorn, Yang, Mitra, Sravankumar, Korenev, Hinsvark, Rao, Zhang, Rodriguez, Gregerson, Spataru, Roziere, Biron, Tang, Chern, Caucheteux, Nayak, Bi, Marra, McConnell, Keller, Touret, Wu, Wong, Ferrer, Nikolaidis, Allonsius, Song, Pintz, Livshits, Wyatt, Esiobu, Choudhary, Mahajan, Garcia-Olano, Perino, Hupkes, Lakomkin, AlBadawy, Lobanova, Dinan, Smith, Radenovic, Guzmán, Zhang, Synnaeve, Lee, Anderson, Thattai, Nail, Mialon, Pang, Cucurell, Nguyen, Korevaar, Xu, Touvron, Zarov, Ibarra, Kloumann, Misra, Evtimov, Zhang, Copet, Lee, Geffert, Vranes, Park, Mahadeokar, Shah, van~der Linde, Billock, Hong, Lee, Fu, Chi, Huang, Liu, Wang, Yu, Bitton, Spisak, Park, Rocca, Johnstun, Saxe, Jia, Alwala, Prasad, Upasani, Plawiak, Li, Heafield, Stone, El-Arini, Iyer, Malik, Chiu, Bhalla, Lakhotia, Rantala-Yeary, van~der Maaten, Chen, Tan, Jenkins, Martin, Madaan, Malo, Blecher,
  Landzaat, de~Oliveira, Muzzi, Pasupuleti, Singh, Paluri, Kardas, Tsimpoukelli, Oldham, Rita, Pavlova, Kambadur, Lewis, Si, Singh, Hassan, Goyal, Torabi, Bashlykov, Bogoychev, Chatterji, Zhang, Duchenne, Çelebi, Alrassy, Zhang, Li, Vasic, Weng, Bhargava, Dubal, Krishnan, Koura, Xu, He, Dong, Srinivasan, Ganapathy, Calderer, Cabral, Stojnic, Raileanu, Maheswari, Girdhar, Patel, Sauvestre, Polidoro, Sumbaly, Taylor, Silva, Hou, Wang, Hosseini, Chennabasappa, Singh, Bell, Kim, Edunov, Nie, Narang, Raparthy, Shen, Wan, Bhosale, Zhang, Vandenhende, Batra, Whitman, Sootla, Collot, Gururangan, Borodinsky, Herman, Fowler, Sheasha, Georgiou, Scialom, Speckbacher, Mihaylov, Xiao, Karn, Goswami, Gupta, Ramanathan, Kerkez, Gonguet, Do, Vogeti, Albiero, Petrovic, Chu, Xiong, Fu, Meers, Martinet, Wang, Wang, Tan, Xia, Xie, Jia, Wang, Goldschlag, Gaur, Babaei, Wen, Song, Zhang, Li, Mao, Coudert, Yan, Chen, Papakipos, Singh, Srivastava, Jain, Kelsey, Shajnfeld, Gangidi, Victoria, Goldstand, Menon, Sharma, Boesenberg,
  Baevski, Feinstein, Kallet, Sangani, Teo, Yunus, Lupu, Alvarado, Caples, Gu, Ho, Poulton, Ryan, Ramchandani, Dong, Franco, Goyal, Saraf, Chowdhury, Gabriel, Bharambe, Eisenman, Yazdan, James, Maurer, Leonhardi, Huang, Loyd, Paola, Paranjape, Liu, Wu, Ni, Hancock, Wasti, Spence, Stojkovic, Gamido, Montalvo, Parker, Burton, Mejia, Liu, Wang, Kim, Zhou, Hu, Chu, Cai, Tindal, Feichtenhofer, Gao, Civin, Beaty, Kreymer, Li, Adkins, Xu, Testuggine, David, Parikh, Liskovich, Foss, Wang, Le, Holland, Dowling, Jamil, Montgomery, Presani, Hahn, Wood, Le, Brinkman, Arcaute, Dunbar, Smothers, Sun, Kreuk, Tian, Kokkinos, Ozgenel, Caggioni, Kanayet, Seide, Florez, Schwarz, Badeer, Swee, Halpern, Herman, Sizov, Guangyi, Zhang, Lakshminarayanan, Inan, Shojanazeri, Zou, Wang, Zha, Habeeb, Rudolph, Suk, Aspegren, Goldman, Zhan, Damlaj, Molybog, Tufanov, Leontiadis, Veliche, Gat, Weissman, Geboski, Kohli, Lam, Asher, Gaya, Marcus, Tang, Chan, Zhen, Reizenstein, Teboul, Zhong, Jin, Yang, Cummings, Carvill, Shepard, McPhie,
  Torres, Ginsburg, Wang, Wu, U, Saxena, Khandelwal, Zand, Matosich, Veeraraghavan, Michelena, Li, Jagadeesh, Huang, Chawla, Huang, Chen, Garg, A, Silva, Bell, Zhang, Guo, Yu, Moshkovich, Wehrstedt, Khabsa, Avalani, Bhatt, Mankus, Hasson, Lennie, Reso, Groshev, Naumov, Lathi, Keneally, Liu, Seltzer, Valko, Restrepo, Patel, Vyatskov, Samvelyan, Clark, Macey, Wang, Hermoso, Metanat, Rastegari, Bansal, Santhanam, Parks, White, Bawa, Singhal, Egebo, Usunier, Mehta, Laptev, Dong, Cheng, Chernoguz, Hart, Salpekar, Kalinli, Kent, Parekh, Saab, Balaji, Rittner, Bontrager, Roux, Dollar, Zvyagina, Ratanchandani, Yuvraj, Liang, Alao, Rodriguez, Ayub, Murthy, Nayani, Mitra, Parthasarathy, Li, Hogan, Battey, Wang, Howes, Rinott, Mehta, Siby, Bondu, Datta, Chugh, Hunt, Dhillon, Sidorov, Pan, Mahajan, Verma, Yamamoto, Ramaswamy, Lindsay, Lindsay, Feng, Lin, Zha, Patil, Shankar, Zhang, Zhang, Wang, Agarwal, Sajuyigbe, Chintala, Max, Chen, Kehoe, Satterfield, Govindaprasad, Gupta, Deng, Cho, Virk, Subramanian, Choudhury,
  Goldman, Remez, Glaser, Best, Koehler, Robinson, Li, Zhang, Matthews, Chou, Shaked, Vontimitta, Ajayi, Montanez, Mohan, Kumar, Mangla, Ionescu, Poenaru, Mihailescu, Ivanov, Li, Wang, Jiang, Bouaziz, Constable, Tang, Wu, Wang, Wu, Gao, Kleinman, Chen, Hu, Jia, Qi, Li, Zhang, Zhang, Adi, Nam, Yu, Wang, Zhao, Hao, Qian, Li, He, Rait, DeVito, Rosnbrick, Wen, Yang, Zhao, and Ma]{grattafiori2024llama3herdmodels}
Grattafiori, A., Dubey, A., Jauhri, A., Pandey, A., Kadian, A., Al-Dahle, A., Letman, A., Mathur, A., Schelten, A., Vaughan, A., Yang, A., Fan, A., Goyal, A., Hartshorn, A., Yang, A., Mitra, A., Sravankumar, A., Korenev, A., Hinsvark, A., Rao, A., Zhang, A., Rodriguez, A., Gregerson, A., Spataru, A., Roziere, B., Biron, B., Tang, B., Chern, B., Caucheteux, C., Nayak, C., Bi, C., Marra, C., McConnell, C., Keller, C., Touret, C., Wu, C., Wong, C., Ferrer, C.~C., Nikolaidis, C., Allonsius, D., Song, D., Pintz, D., Livshits, D., Wyatt, D., Esiobu, D., Choudhary, D., Mahajan, D., Garcia-Olano, D., Perino, D., Hupkes, D., Lakomkin, E., AlBadawy, E., Lobanova, E., Dinan, E., Smith, E.~M., Radenovic, F., Guzmán, F., Zhang, F., Synnaeve, G., Lee, G., Anderson, G.~L., Thattai, G., Nail, G., Mialon, G., Pang, G., Cucurell, G., Nguyen, H., Korevaar, H., Xu, H., Touvron, H., Zarov, I., Ibarra, I.~A., Kloumann, I., Misra, I., Evtimov, I., Zhang, J., Copet, J., Lee, J., Geffert, J., Vranes, J., Park, J., Mahadeokar, J.,
  Shah, J., van~der Linde, J., Billock, J., Hong, J., Lee, J., Fu, J., Chi, J., Huang, J., Liu, J., Wang, J., Yu, J., Bitton, J., Spisak, J., Park, J., Rocca, J., Johnstun, J., Saxe, J., Jia, J., Alwala, K.~V., Prasad, K., Upasani, K., Plawiak, K., Li, K., Heafield, K., Stone, K., El-Arini, K., Iyer, K., Malik, K., Chiu, K., Bhalla, K., Lakhotia, K., Rantala-Yeary, L., van~der Maaten, L., Chen, L., Tan, L., Jenkins, L., Martin, L., Madaan, L., Malo, L., Blecher, L., Landzaat, L., de~Oliveira, L., Muzzi, M., Pasupuleti, M., Singh, M., Paluri, M., Kardas, M., Tsimpoukelli, M., Oldham, M., Rita, M., Pavlova, M., Kambadur, M., Lewis, M., Si, M., Singh, M.~K., Hassan, M., Goyal, N., Torabi, N., Bashlykov, N., Bogoychev, N., Chatterji, N., Zhang, N., Duchenne, O., Çelebi, O., Alrassy, P., Zhang, P., Li, P., Vasic, P., Weng, P., Bhargava, P., Dubal, P., Krishnan, P., Koura, P.~S., Xu, P., He, Q., Dong, Q., Srinivasan, R., Ganapathy, R., Calderer, R., Cabral, R.~S., Stojnic, R., Raileanu, R., Maheswari, R., Girdhar,
  R., Patel, R., Sauvestre, R., Polidoro, R., Sumbaly, R., Taylor, R., Silva, R., Hou, R., Wang, R., Hosseini, S., Chennabasappa, S., Singh, S., Bell, S., Kim, S.~S., Edunov, S., Nie, S., Narang, S., Raparthy, S., Shen, S., Wan, S., Bhosale, S., Zhang, S., Vandenhende, S., Batra, S., Whitman, S., Sootla, S., Collot, S., Gururangan, S., Borodinsky, S., Herman, T., Fowler, T., Sheasha, T., Georgiou, T., Scialom, T., Speckbacher, T., Mihaylov, T., Xiao, T., Karn, U., Goswami, V., Gupta, V., Ramanathan, V., Kerkez, V., Gonguet, V., Do, V., Vogeti, V., Albiero, V., Petrovic, V., Chu, W., Xiong, W., Fu, W., Meers, W., Martinet, X., Wang, X., Wang, X., Tan, X.~E., Xia, X., Xie, X., Jia, X., Wang, X., Goldschlag, Y., Gaur, Y., Babaei, Y., Wen, Y., Song, Y., Zhang, Y., Li, Y., Mao, Y., Coudert, Z.~D., Yan, Z., Chen, Z., Papakipos, Z., Singh, A., Srivastava, A., Jain, A., Kelsey, A., Shajnfeld, A., Gangidi, A., Victoria, A., Goldstand, A., Menon, A., Sharma, A., Boesenberg, A., Baevski, A., Feinstein, A., Kallet, A.,
  Sangani, A., Teo, A., Yunus, A., Lupu, A., Alvarado, A., Caples, A., Gu, A., Ho, A., Poulton, A., Ryan, A., Ramchandani, A., Dong, A., Franco, A., Goyal, A., Saraf, A., Chowdhury, A., Gabriel, A., Bharambe, A., Eisenman, A., Yazdan, A., James, B., Maurer, B., Leonhardi, B., Huang, B., Loyd, B., Paola, B.~D., Paranjape, B., Liu, B., Wu, B., Ni, B., Hancock, B., Wasti, B., Spence, B., Stojkovic, B., Gamido, B., Montalvo, B., Parker, C., Burton, C., Mejia, C., Liu, C., Wang, C., Kim, C., Zhou, C., Hu, C., Chu, C.-H., Cai, C., Tindal, C., Feichtenhofer, C., Gao, C., Civin, D., Beaty, D., Kreymer, D., Li, D., Adkins, D., Xu, D., Testuggine, D., David, D., Parikh, D., Liskovich, D., Foss, D., Wang, D., Le, D., Holland, D., Dowling, E., Jamil, E., Montgomery, E., Presani, E., Hahn, E., Wood, E., Le, E.-T., Brinkman, E., Arcaute, E., Dunbar, E., Smothers, E., Sun, F., Kreuk, F., Tian, F., Kokkinos, F., Ozgenel, F., Caggioni, F., Kanayet, F., Seide, F., Florez, G.~M., Schwarz, G., Badeer, G., Swee, G., Halpern, G.,
  Herman, G., Sizov, G., Guangyi, Zhang, Lakshminarayanan, G., Inan, H., Shojanazeri, H., Zou, H., Wang, H., Zha, H., Habeeb, H., Rudolph, H., Suk, H., Aspegren, H., Goldman, H., Zhan, H., Damlaj, I., Molybog, I., Tufanov, I., Leontiadis, I., Veliche, I.-E., Gat, I., Weissman, J., Geboski, J., Kohli, J., Lam, J., Asher, J., Gaya, J.-B., Marcus, J., Tang, J., Chan, J., Zhen, J., Reizenstein, J., Teboul, J., Zhong, J., Jin, J., Yang, J., Cummings, J., Carvill, J., Shepard, J., McPhie, J., Torres, J., Ginsburg, J., Wang, J., Wu, K., U, K.~H., Saxena, K., Khandelwal, K., Zand, K., Matosich, K., Veeraraghavan, K., Michelena, K., Li, K., Jagadeesh, K., Huang, K., Chawla, K., Huang, K., Chen, L., Garg, L., A, L., Silva, L., Bell, L., Zhang, L., Guo, L., Yu, L., Moshkovich, L., Wehrstedt, L., Khabsa, M., Avalani, M., Bhatt, M., Mankus, M., Hasson, M., Lennie, M., Reso, M., Groshev, M., Naumov, M., Lathi, M., Keneally, M., Liu, M., Seltzer, M.~L., Valko, M., Restrepo, M., Patel, M., Vyatskov, M., Samvelyan, M., Clark,
  M., Macey, M., Wang, M., Hermoso, M.~J., Metanat, M., Rastegari, M., Bansal, M., Santhanam, N., Parks, N., White, N., Bawa, N., Singhal, N., Egebo, N., Usunier, N., Mehta, N., Laptev, N.~P., Dong, N., Cheng, N., Chernoguz, O., Hart, O., Salpekar, O., Kalinli, O., Kent, P., Parekh, P., Saab, P., Balaji, P., Rittner, P., Bontrager, P., Roux, P., Dollar, P., Zvyagina, P., Ratanchandani, P., Yuvraj, P., Liang, Q., Alao, R., Rodriguez, R., Ayub, R., Murthy, R., Nayani, R., Mitra, R., Parthasarathy, R., Li, R., Hogan, R., Battey, R., Wang, R., Howes, R., Rinott, R., Mehta, S., Siby, S., Bondu, S.~J., Datta, S., Chugh, S., Hunt, S., Dhillon, S., Sidorov, S., Pan, S., Mahajan, S., Verma, S., Yamamoto, S., Ramaswamy, S., Lindsay, S., Lindsay, S., Feng, S., Lin, S., Zha, S.~C., Patil, S., Shankar, S., Zhang, S., Zhang, S., Wang, S., Agarwal, S., Sajuyigbe, S., Chintala, S., Max, S., Chen, S., Kehoe, S., Satterfield, S., Govindaprasad, S., Gupta, S., Deng, S., Cho, S., Virk, S., Subramanian, S., Choudhury, S.,
  Goldman, S., Remez, T., Glaser, T., Best, T., Koehler, T., Robinson, T., Li, T., Zhang, T., Matthews, T., Chou, T., Shaked, T., Vontimitta, V., Ajayi, V., Montanez, V., Mohan, V., Kumar, V.~S., Mangla, V., Ionescu, V., Poenaru, V., Mihailescu, V.~T., Ivanov, V., Li, W., Wang, W., Jiang, W., Bouaziz, W., Constable, W., Tang, X., Wu, X., Wang, X., Wu, X., Gao, X., Kleinman, Y., Chen, Y., Hu, Y., Jia, Y., Qi, Y., Li, Y., Zhang, Y., Zhang, Y., Adi, Y., Nam, Y., Yu, Wang, Zhao, Y., Hao, Y., Qian, Y., Li, Y., He, Y., Rait, Z., DeVito, Z., Rosnbrick, Z., Wen, Z., Yang, Z., Zhao, Z., and Ma, Z.
\newblock The llama 3 herd of models, 2024.
\newblock URL \url{https://arxiv.org/abs/2407.21783}.

\bibitem[Hendrycks et~al.(2021)Hendrycks, Burns, Kadavath, Arora, Basart, Tang, Song, and Steinhardt]{NEURIPS_DATASETS_AND_BENCHMARKS2021_be83ab3e}
Hendrycks, D., Burns, C., Kadavath, S., Arora, A., Basart, S., Tang, E., Song, D., and Steinhardt, J.
\newblock Measuring mathematical problem solving with the math dataset.
\newblock In Vanschoren, J. and Yeung, S. (eds.), \emph{Proceedings of the Neural Information Processing Systems Track on Datasets and Benchmarks}, volume~1, 2021.
\newblock URL \url{https://datasets-benchmarks-proceedings.neurips.cc/paper_files/paper/2021/file/be83ab3ecd0db773eb2dc1b0a17836a1-Paper-round2.pdf}.

\bibitem[Hooper et~al.(2024)Hooper, Kim, Mohammadzadeh, Mahoney, Shao, Keutzer, and Gholami]{hooper2024kvquant}
Hooper, C. R.~C., Kim, S., Mohammadzadeh, H., Mahoney, M.~W., Shao, S., Keutzer, K., and Gholami, A.
\newblock {KVQ}uant: Towards 10 million context length {LLM} inference with {KV} cache quantization.
\newblock In \emph{The Thirty-eighth Annual Conference on Neural Information Processing Systems}, 2024.
\newblock URL \url{https://openreview.net/forum?id=0LXotew9Du}.

\bibitem[Hui et~al.(2024)Hui, Yang, Cui, Yang, Liu, Zhang, Liu, Zhang, Yu, Lu, Dang, Fan, Zhang, Yang, Men, Huang, Zheng, Miao, Quan, Feng, Ren, Ren, Zhou, and Lin]{hui2024qwen25codertechnicalreport}
Hui, B., Yang, J., Cui, Z., Yang, J., Liu, D., Zhang, L., Liu, T., Zhang, J., Yu, B., Lu, K., Dang, K., Fan, Y., Zhang, Y., Yang, A., Men, R., Huang, F., Zheng, B., Miao, Y., Quan, S., Feng, Y., Ren, X., Ren, X., Zhou, J., and Lin, J.
\newblock Qwen2.5-coder technical report, 2024.
\newblock URL \url{https://arxiv.org/abs/2409.12186}.

\bibitem[Jia et~al.(2024)Jia, Xie, Lu, Wang, Feng, Zhang, Sun, Lin, Zhang, Liu, and Tao]{jia2024sdpbit}
Jia, J., Xie, C., Lu, H., Wang, D., Feng, H., Zhang, C., Sun, B., Lin, H., Zhang, Z., Liu, X., and Tao, D.
\newblock {SDP}4bit: Toward 4-bit communication quantization in sharded data parallelism for {LLM} training.
\newblock In \emph{The Thirty-eighth Annual Conference on Neural Information Processing Systems}, 2024.
\newblock URL \url{https://openreview.net/forum?id=PEEqnXlSCk}.

\bibitem[Kim et~al.(2024)Kim, Hooper, Gholami, Dong, Li, Shen, Mahoney, and Keutzer]{pmlr-v235-kim24f}
Kim, S., Hooper, C. R.~C., Gholami, A., Dong, Z., Li, X., Shen, S., Mahoney, M.~W., and Keutzer, K.
\newblock {S}queeze{LLM}: Dense-and-sparse quantization.
\newblock In Salakhutdinov, R., Kolter, Z., Heller, K., Weller, A., Oliver, N., Scarlett, J., and Berkenkamp, F. (eds.), \emph{Proceedings of the 41st International Conference on Machine Learning}, volume 235 of \emph{Proceedings of Machine Learning Research}, pp.\  23901--23923. PMLR, 21--27 Jul 2024.
\newblock URL \url{https://proceedings.mlr.press/v235/kim24f.html}.

\bibitem[Kumar et~al.(2024)Kumar, Ankner, Spector, Bordelon, Muennighoff, Paul, Pehlevan, Ré, and Raghunathan]{kumar2024scalinglawsprecision}
Kumar, T., Ankner, Z., Spector, B.~F., Bordelon, B., Muennighoff, N., Paul, M., Pehlevan, C., Ré, C., and Raghunathan, A.
\newblock Scaling laws for precision, 2024.
\newblock URL \url{https://arxiv.org/abs/2411.04330}.

\bibitem[Lambert et~al.(2024)Lambert, Morrison, Pyatkin, Huang, Ivison, Brahman, Miranda, Liu, Dziri, Lyu, Gu, Malik, Graf, Hwang, Yang, Bras, Tafjord, Wilhelm, Soldaini, Smith, Wang, Dasigi, and Hajishirzi]{lambert2024tulu3}
Lambert, N., Morrison, J., Pyatkin, V., Huang, S., Ivison, H., Brahman, F., Miranda, L. J.~V., Liu, A., Dziri, N., Lyu, S., Gu, Y., Malik, S., Graf, V., Hwang, J.~D., Yang, J., Bras, R.~L., Tafjord, O., Wilhelm, C., Soldaini, L., Smith, N.~A., Wang, Y., Dasigi, P., and Hajishirzi, H.
\newblock Tülu 3: Pushing frontiers in open language model post-training.
\newblock 2024.

\bibitem[LeCun et~al.(1989)LeCun, Denker, and Solla]{NIPS1989_6c9882bb}
LeCun, Y., Denker, J., and Solla, S.
\newblock Optimal brain damage.
\newblock In Touretzky, D. (ed.), \emph{Advances in Neural Information Processing Systems}, volume~2. Morgan-Kaufmann, 1989.
\newblock URL \url{https://proceedings.neurips.cc/paper_files/paper/1989/file/6c9882bbac1c7093bd25041881277658-Paper.pdf}.

\bibitem[Li et~al.(2023{\natexlab{a}})Li, Chen, and Zhu]{NEURIPS2023_3122aaa2}
Li, B., Chen, J., and Zhu, J.
\newblock Memory efficient optimizers with 4-bit states.
\newblock In Oh, A., Naumann, T., Globerson, A., Saenko, K., Hardt, M., and Levine, S. (eds.), \emph{Advances in Neural Information Processing Systems}, volume~36, pp.\  15136--15171. Curran Associates, Inc., 2023{\natexlab{a}}.
\newblock URL \url{https://proceedings.neurips.cc/paper_files/paper/2023/file/3122aaa22b2fe83f9cead1a696f65ceb-Paper-Conference.pdf}.

\bibitem[Li \& Yuan(2017)Li and Yuan]{li2017convergence}
Li, Y. and Yuan, Y.
\newblock Convergence analysis of two-layer neural networks with relu activation.
\newblock In Guyon, I., Luxburg, U.~V., Bengio, S., Wallach, H., Fergus, R., Vishwanathan, S., and Garnett, R. (eds.), \emph{Advances in Neural Information Processing Systems}, volume~30. Curran Associates, Inc., 2017.
\newblock URL \url{https://proceedings.neurips.cc/paper_files/paper/2017/file/a96b65a721e561e1e3de768ac819ffbb-Paper.pdf}.

\bibitem[Li et~al.(2023{\natexlab{b}})Li, Yu, Zhang, Liang, He, Chen, and Zhao]{li2023losparsestructuredcompressionlarge}
Li, Y., Yu, Y., Zhang, Q., Liang, C., He, P., Chen, W., and Zhao, T.
\newblock Losparse: Structured compression of large language models based on low-rank and sparse approximation, 2023{\natexlab{b}}.
\newblock URL \url{https://arxiv.org/abs/2306.11222}.

\bibitem[Li et~al.(2024)Li, Yu, Liang, Karampatziakis, He, Chen, and Zhao]{li2024loftq}
Li, Y., Yu, Y., Liang, C., Karampatziakis, N., He, P., Chen, W., and Zhao, T.
\newblock Loftq: Lo{RA}-fine-tuning-aware quantization for large language models.
\newblock In \emph{The Twelfth International Conference on Learning Representations}, 2024.
\newblock URL \url{https://openreview.net/forum?id=LzPWWPAdY4}.

\bibitem[Lin et~al.(2024{\natexlab{a}})Lin, Xu, Wu, Cui, Zhang, Mou, Song, Sun, and Wei]{lin2024duquant}
Lin, H., Xu, H., Wu, Y., Cui, J., Zhang, Y., Mou, L., Song, L., Sun, Z., and Wei, Y.
\newblock Duquant: Distributing outliers via dual transformation makes stronger quantized {LLM}s.
\newblock In \emph{The Thirty-eighth Annual Conference on Neural Information Processing Systems}, 2024{\natexlab{a}}.
\newblock URL \url{https://openreview.net/forum?id=mp8u2Pcmqz}.

\bibitem[Lin et~al.(2024{\natexlab{b}})Lin, Tang, Tang, Yang, Chen, Wang, Xiao, Dang, Gan, and Han]{MLSYS2024_42a452cb}
Lin, J., Tang, J., Tang, H., Yang, S., Chen, W.-M., Wang, W.-C., Xiao, G., Dang, X., Gan, C., and Han, S.
\newblock Awq: Activation-aware weight quantization for on-device llm compression and acceleration.
\newblock In Gibbons, P., Pekhimenko, G., and Sa, C.~D. (eds.), \emph{Proceedings of Machine Learning and Systems}, volume~6, pp.\  87--100, 2024{\natexlab{b}}.
\newblock URL \url{https://proceedings.mlsys.org/paper_files/paper/2024/file/42a452cbafa9dd64e9ba4aa95cc1ef21-Paper-Conference.pdf}.

\bibitem[Lin* et~al.(2024)Lin*, Tang*, Yang*, Zhang, Xiao, Gan, and Han]{lin2024qserve}
Lin*, Y., Tang*, H., Yang*, S., Zhang, Z., Xiao, G., Gan, C., and Han, S.
\newblock Qserve: W4a8kv4 quantization and system co-design for efficient llm serving.
\newblock \emph{arXiv preprint arXiv:2405.04532}, 2024.

\bibitem[Liu et~al.(2023)Liu, Oguz, Zhao, Chang, Stock, Mehdad, Shi, Krishnamoorthi, and Chandra]{liu2023llmqatdatafreequantizationaware}
Liu, Z., Oguz, B., Zhao, C., Chang, E., Stock, P., Mehdad, Y., Shi, Y., Krishnamoorthi, R., and Chandra, V.
\newblock Llm-qat: Data-free quantization aware training for large language models, 2023.
\newblock URL \url{https://arxiv.org/abs/2305.17888}.

\bibitem[Liu et~al.(2024{\natexlab{a}})Liu, Oguz, Zhao, Chang, Stock, Mehdad, Shi, Krishnamoorthi, and Chandra]{liu-etal-2024-llm}
Liu, Z., Oguz, B., Zhao, C., Chang, E., Stock, P., Mehdad, Y., Shi, Y., Krishnamoorthi, R., and Chandra, V.
\newblock {LLM}-{QAT}: Data-free quantization aware training for large language models.
\newblock In Ku, L.-W., Martins, A., and Srikumar, V. (eds.), \emph{Findings of the Association for Computational Linguistics: ACL 2024}, pp.\  467--484, Bangkok, Thailand, August 2024{\natexlab{a}}. Association for Computational Linguistics.
\newblock \doi{10.18653/v1/2024.findings-acl.26}.
\newblock URL \url{https://aclanthology.org/2024.findings-acl.26/}.

\bibitem[Liu et~al.(2024{\natexlab{b}})Liu, Zhao, Fedorov, Soran, Choudhary, Krishnamoorthi, Chandra, Tian, and Blankevoort]{liu2024spinquantllmquantizationlearned}
Liu, Z., Zhao, C., Fedorov, I., Soran, B., Choudhary, D., Krishnamoorthi, R., Chandra, V., Tian, Y., and Blankevoort, T.
\newblock Spinquant: Llm quantization with learned rotations, 2024{\natexlab{b}}.
\newblock URL \url{https://arxiv.org/abs/2405.16406}.

\bibitem[Ma et~al.(2024)Ma, Li, Zheng, Ling, Xiao, Wang, Wen, Chao, and Ji]{ma2024affinequant}
Ma, Y., Li, H., Zheng, X., Ling, F., Xiao, X., Wang, R., Wen, S., Chao, F., and Ji, R.
\newblock Affinequant: Affine transformation quantization for large language models.
\newblock In \emph{The Twelfth International Conference on Learning Representations}, 2024.
\newblock URL \url{https://openreview.net/forum?id=of2rhALq8l}.

\bibitem[Merity et~al.(2017)Merity, Xiong, Bradbury, and Socher]{merity2017pointer}
Merity, S., Xiong, C., Bradbury, J., and Socher, R.
\newblock Pointer sentinel mixture models.
\newblock In \emph{International Conference on Learning Representations}, 2017.
\newblock URL \url{https://openreview.net/forum?id=Byj72udxe}.

\bibitem[Ouyang et~al.(2022)Ouyang, Wu, Jiang, Almeida, Wainwright, Mishkin, Zhang, Agarwal, Slama, Ray, Schulman, Hilton, Kelton, Miller, Simens, Askell, Welinder, Christiano, Leike, and Lowe]{ouyang2022traininglanguagemodelsfollow}
Ouyang, L., Wu, J., Jiang, X., Almeida, D., Wainwright, C.~L., Mishkin, P., Zhang, C., Agarwal, S., Slama, K., Ray, A., Schulman, J., Hilton, J., Kelton, F., Miller, L., Simens, M., Askell, A., Welinder, P., Christiano, P., Leike, J., and Lowe, R.
\newblock Training language models to follow instructions with human feedback, 2022.
\newblock URL \url{https://arxiv.org/abs/2203.02155}.

\bibitem[Shao et~al.(2024{\natexlab{a}})Shao, Chen, Zhang, Xu, Zhao, Li, Zhang, Gao, Qiao, and Luo]{shao2024omniquant}
Shao, W., Chen, M., Zhang, Z., Xu, P., Zhao, L., Li, Z., Zhang, K., Gao, P., Qiao, Y., and Luo, P.
\newblock Omniquant: Omnidirectionally calibrated quantization for large language models.
\newblock In \emph{The Twelfth International Conference on Learning Representations}, 2024{\natexlab{a}}.
\newblock URL \url{https://openreview.net/forum?id=8Wuvhh0LYW}.

\bibitem[Shao et~al.(2024{\natexlab{b}})Shao, Liang, Ling, Yan, Liu, Chen, Yan, Zhang, Qin, Magno, Yang, Lei, Wang, Guo, Shao, and Tang]{shao2024gwqgradientawareweightquantization}
Shao, Y., Liang, S., Ling, Z., Yan, M., Liu, H., Chen, S., Yan, Z., Zhang, C., Qin, H., Magno, M., Yang, Y., Lei, Z., Wang, Y., Guo, J., Shao, L., and Tang, H.
\newblock Gwq: Gradient-aware weight quantization for large language models, 2024{\natexlab{b}}.
\newblock URL \url{https://arxiv.org/abs/2411.00850}.

\bibitem[Touvron et~al.(2023)Touvron, Martin, Stone, Albert, Almahairi, Babaei, Bashlykov, Batra, Bhargava, Bhosale, Bikel, Blecher, Ferrer, Chen, Cucurull, Esiobu, Fernandes, Fu, Fu, Fuller, Gao, Goswami, Goyal, Hartshorn, Hosseini, Hou, Inan, Kardas, Kerkez, Khabsa, Kloumann, Korenev, Koura, Lachaux, Lavril, Lee, Liskovich, Lu, Mao, Martinet, Mihaylov, Mishra, Molybog, Nie, Poulton, Reizenstein, Rungta, Saladi, Schelten, Silva, Smith, Subramanian, Tan, Tang, Taylor, Williams, Kuan, Xu, Yan, Zarov, Zhang, Fan, Kambadur, Narang, Rodriguez, Stojnic, Edunov, and Scialom]{touvron2023llama2openfoundation}
Touvron, H., Martin, L., Stone, K., Albert, P., Almahairi, A., Babaei, Y., Bashlykov, N., Batra, S., Bhargava, P., Bhosale, S., Bikel, D., Blecher, L., Ferrer, C.~C., Chen, M., Cucurull, G., Esiobu, D., Fernandes, J., Fu, J., Fu, W., Fuller, B., Gao, C., Goswami, V., Goyal, N., Hartshorn, A., Hosseini, S., Hou, R., Inan, H., Kardas, M., Kerkez, V., Khabsa, M., Kloumann, I., Korenev, A., Koura, P.~S., Lachaux, M.-A., Lavril, T., Lee, J., Liskovich, D., Lu, Y., Mao, Y., Martinet, X., Mihaylov, T., Mishra, P., Molybog, I., Nie, Y., Poulton, A., Reizenstein, J., Rungta, R., Saladi, K., Schelten, A., Silva, R., Smith, E.~M., Subramanian, R., Tan, X.~E., Tang, B., Taylor, R., Williams, A., Kuan, J.~X., Xu, P., Yan, Z., Zarov, I., Zhang, Y., Fan, A., Kambadur, M., Narang, S., Rodriguez, A., Stojnic, R., Edunov, S., and Scialom, T.
\newblock Llama 2: Open foundation and fine-tuned chat models, 2023.
\newblock URL \url{https://arxiv.org/abs/2307.09288}.

\bibitem[Tseng et~al.(2024{\natexlab{a}})Tseng, Chee, Sun, Kuleshov, and Sa]{tseng2024quip}
Tseng, A., Chee, J., Sun, Q., Kuleshov, V., and Sa, C.~D.
\newblock Qu{IP}\${\textbackslash}\#\$: Even better {LLM} quantization with hadamard incoherence and lattice codebooks.
\newblock In \emph{Forty-first International Conference on Machine Learning}, 2024{\natexlab{a}}.
\newblock URL \url{https://openreview.net/forum?id=9BrydUVcoe}.

\bibitem[Tseng et~al.(2024{\natexlab{b}})Tseng, Sun, Hou, and Sa]{tseng2024qtip}
Tseng, A., Sun, Q., Hou, D., and Sa, C.~D.
\newblock {QTIP}: Quantization with trellises and incoherence processing.
\newblock In \emph{The Thirty-eighth Annual Conference on Neural Information Processing Systems}, 2024{\natexlab{b}}.
\newblock URL \url{https://openreview.net/forum?id=7sdkLVuYCU}.

\bibitem[Wang et~al.(2024)Wang, Zhou, Song, Mao, Ma, Wang, Xia, and Wei]{wang20241bitaiinfra11}
Wang, J., Zhou, H., Song, T., Mao, S., Ma, S., Wang, H., Xia, Y., and Wei, F.
\newblock 1-bit ai infra: Part 1.1, fast and lossless bitnet b1.58 inference on cpus, 2024.
\newblock URL \url{https://arxiv.org/abs/2410.16144}.

\bibitem[Weber et~al.(2024)Weber, Fu, Anthony, Oren, Adams, Alexandrov, Lyu, Nguyen, Yao, Adams, Athiwaratkun, Chalamala, Chen, Ryabinin, Dao, Liang, Ré, Rish, and Zhang]{weber2024redpajama}
Weber, M., Fu, D.~Y., Anthony, Q., Oren, Y., Adams, S., Alexandrov, A., Lyu, X., Nguyen, H., Yao, X., Adams, V., Athiwaratkun, B., Chalamala, R., Chen, K., Ryabinin, M., Dao, T., Liang, P., Ré, C., Rish, I., and Zhang, C.
\newblock Redpajama: an open dataset for training large language models.
\newblock \emph{NeurIPS Datasets and Benchmarks Track}, 2024.

\bibitem[Wei et~al.(2022)Wei, Wang, Schuurmans, Bosma, ichter, Xia, Chi, Le, and Zhou]{NEURIPS2022_9d560961}
Wei, J., Wang, X., Schuurmans, D., Bosma, M., ichter, b., Xia, F., Chi, E., Le, Q.~V., and Zhou, D.
\newblock Chain-of-thought prompting elicits reasoning in large language models.
\newblock In Koyejo, S., Mohamed, S., Agarwal, A., Belgrave, D., Cho, K., and Oh, A. (eds.), \emph{Advances in Neural Information Processing Systems}, volume~35, pp.\  24824--24837. Curran Associates, Inc., 2022.
\newblock URL \url{https://proceedings.neurips.cc/paper_files/paper/2022/file/9d5609613524ecf4f15af0f7b31abca4-Paper-Conference.pdf}.

\bibitem[Xia et~al.(2024)Xia, Fu, Zhang, Jiang, and CUI]{xia2024efficient}
Xia, Y., Fu, F., Zhang, W., Jiang, J., and CUI, B.
\newblock Efficient multi-task {LLM} quantization and serving for multiple lo{RA} adapters.
\newblock In \emph{The Thirty-eighth Annual Conference on Neural Information Processing Systems}, 2024.
\newblock URL \url{https://openreview.net/forum?id=HfpV6u0kbX}.

\bibitem[Xiao et~al.(2023)Xiao, Lin, Seznec, Wu, Demouth, and Han]{pmlr-v202-xiao23c}
Xiao, G., Lin, J., Seznec, M., Wu, H., Demouth, J., and Han, S.
\newblock {S}mooth{Q}uant: Accurate and efficient post-training quantization for large language models.
\newblock In Krause, A., Brunskill, E., Cho, K., Engelhardt, B., Sabato, S., and Scarlett, J. (eds.), \emph{Proceedings of the 40th International Conference on Machine Learning}, volume 202 of \emph{Proceedings of Machine Learning Research}, pp.\  38087--38099. PMLR, 23--29 Jul 2023.
\newblock URL \url{https://proceedings.mlr.press/v202/xiao23c.html}.

\bibitem[Zhang et~al.(2024)Zhang, Huang, Zhang, Wei, Zhu, and Chen]{zhang2024sageattention2}
Zhang, J., Huang, H., Zhang, P., Wei, J., Zhu, J., and Chen, J.
\newblock Sageattention2: Efficient attention with thorough outlier smoothing and per-thread int4 quantization, 2024.
\newblock URL \url{https://arxiv.org/abs/2411.10958}.

\bibitem[Zhang et~al.(2025)Zhang, Wei, Zhang, Zhu, and Chen]{zhang2025sageattention}
Zhang, J., Wei, J., Zhang, P., Zhu, J., and Chen, J.
\newblock Sageattention: Accurate 8-bit attention for plug-and-play inference acceleration.
\newblock In \emph{International Conference on Learning Representations (ICLR)}, 2025.

\bibitem[Zhang et~al.(2022)Zhang, Roller, Goyal, Artetxe, Chen, Chen, Dewan, Diab, Li, Lin, Mihaylov, Ott, Shleifer, Shuster, Simig, Koura, Sridhar, Wang, and Zettlemoyer]{zhang2022optopenpretrainedtransformer}
Zhang, S., Roller, S., Goyal, N., Artetxe, M., Chen, M., Chen, S., Dewan, C., Diab, M., Li, X., Lin, X.~V., Mihaylov, T., Ott, M., Shleifer, S., Shuster, K., Simig, D., Koura, P.~S., Sridhar, A., Wang, T., and Zettlemoyer, L.
\newblock Opt: Open pre-trained transformer language models, 2022.
\newblock URL \url{https://arxiv.org/abs/2205.01068}.

\end{thebibliography}
